\pdfoutput=1

\documentclass[11pt]{article}

\usepackage{acl}

\usepackage{times}
\usepackage{latexsym}

\usepackage[T1]{fontenc}

\usepackage[utf8]{inputenc}

\usepackage{microtype}

\usepackage{inconsolata}

\usepackage{enumitem}
\usepackage{refstyle}
\usepackage{xspace}
\usepackage{colortbl}

\usepackage{booktabs}
\usepackage{multirow} 
\usepackage{soul}
\usepackage{tabularx}
\usepackage{array}         
\usepackage{siunitx}       
\usepackage{svg}
\usepackage{amsfonts}
\usepackage{marvosym}

\usepackage{amsmath,amsfonts,amssymb}
\usepackage{mathrsfs}
\usepackage{pifont}

\usepackage{cleveref}
\crefformat{section}{\S#2#1#3}
\crefformat{subsection}{\S#2#1#3}
\crefformat{subsubsection}{\S#2#1#3}
\crefrangeformat{section}{\S#3#1#4 to~\S#5#2#6}
\crefmultiformat{section}{\S#2#1#3}{ and~\S#2#1#3}{, #2#1#3}{ and~#2#1#3}
\Crefformat{figure}{#2Fig.~#1#3}
\Crefmultiformat{figure}{Figs.~#2#1#3}{ and~#2#1#3}{, #2#1#3}{ and~#2#1#3}
\Crefformat{table}{#2Tab.~#1#3}
\Crefmultiformat{table}{Tabs.~#2#1#3}{ and~#2#1#3}{, #2#1#3}{ and~#2#1#3}
\Crefformat{appendix}{#2Appx.~\S#1#3}
\crefformat{algorithm}{Alg.~#2#1#3}
\Crefformat{equation}{#2Eq.~#1#3}

\newcommand{\stitle}[1]{\vspace{1ex} \noindent{\bf #1}}

\usepackage{caption}
\usepackage{subcaption}
\usepackage{tcolorbox}
\usepackage{setspace}
\usepackage{listings}
\usepackage{booktabs} 
\usepackage{xcolor}
\usepackage{colortbl} 
\usepackage{makecell}
\usepackage{amssymb}
\usepackage{tablefootnote}

\definecolor{darkred}{rgb}{0.75, 0.0, 0.0}
\definecolor{darkgreen}{rgb}{0.0, 0.45, 0.0}

\definecolor{mint}{rgb}{0.62, 0.89, 0.75}
\definecolor{lightrose}{rgb}{0.93, 0.26, 0.22}

\colorlet{mint5}{mint!5}
\colorlet{mint10}{mint!10}
\colorlet{mint20}{mint!20}
\colorlet{mint30}{mint!30}
\colorlet{mint40}{mint!40}
\colorlet{mint50}{mint!50}
\colorlet{mint60}{mint!60}
\colorlet{mint70}{mint!70}
\colorlet{mint80}{mint!80}
\colorlet{mint90}{mint!90}

\colorlet{rose5}{lightrose!5}
\colorlet{rose10}{lightrose!10}
\colorlet{rose20}{lightrose!20}
\colorlet{rose30}{lightrose!30}
\colorlet{rose40}{lightrose!40}
\colorlet{rose45}{lightrose!45}
\colorlet{rose50}{lightrose!50}
\colorlet{rose60}{lightrose!60}
\colorlet{rose70}{lightrose!70}
\colorlet{rose80}{lightrose!80}
\colorlet{rose90}{lightrose!90}

\usepackage{graphicx}

\newcommand{\smallbold}[1]{\textbf{\scalebox{0.8}{\fontfamily{ppl}\selectfont #1}}}

\definecolor{mygray}{gray}{.9}
\usepackage{arydshln}
\usepackage{wrapfig}

\title{Rethinking Tabular Data Understanding with Large Language Models}

\author{Tianyang Liu \\
  UC San Diego \\
  \texttt{til040@ucsd.edu} \\\And
  Fei Wang\\
  USC \\
  \texttt{fwang598@usc.edu} \\\And
  Muhao Chen\\
  UC Davis\\
  \texttt{muhchen@ucdavis.edu} }

\begin{document}
\maketitle

\begin{abstract}
Large Language Models (LLMs) have shown to be capable of various tasks, yet their capability in interpreting and reasoning over tabular data remains an underexplored area. In this context, this study investigates from three core perspectives: the robustness of LLMs to structural perturbations in tables, the comparative analysis of textual and symbolic reasoning on tables, and the potential of boosting model performance through the aggregation of multiple reasoning pathways. We discover that structural variance of tables presenting the same content reveals a notable performance decline, particularly in symbolic reasoning tasks. This prompts the proposal of a method for table structure normalization. Moreover, textual reasoning slightly edges out symbolic reasoning, and a detailed error analysis reveals that each exhibits different strengths depending on the specific tasks. Notably, the aggregation of textual and symbolic reasoning pathways, bolstered by a mix self-consistency mechanism, resulted in achieving SOTA performance, with an accuracy of 73.6\% on \textsc{WikiTableQuestions}, representing a substantial advancement over previous existing table processing paradigms of LLMs.
\end{abstract}
\section{Introduction}\label{sec:intro}


Large Language Models (LLMs; \citealt{brown2020language, PaLM, zhang2022opt, openai2022codex,openai2023chatgpt,openai2023gpt4, touvron2023llama, touvron2023llama2}) have revolutionized the field of NLP, demonstrating an extraordinary ability to understand and reason over rich textual data~\cite{wei2023chainofthought, wang2023selfconsistency, zhou2023leasttomost, kojima2023large, li2023making}. 
On top of LLMs' existing capabilities for NLP,
further bolstering their potential for decision-making by drawing from external knowledge sources remains an exciting research frontier~\cite{nakano2022webgpt, mialon2023augmented, hao2023toolkengpt, jiang2023active}. 
Amongst such knowledge sources,
tabular data serve as a ubiquitous kind due to their expressiveness for relations, properties and statistics, and their being easy to construct by human curators.

\setlength{\fboxsep}{0pt}

\begin{figure}[!t]
    \centering
    \vspace{-0.8em}
    \includegraphics[width=\linewidth]{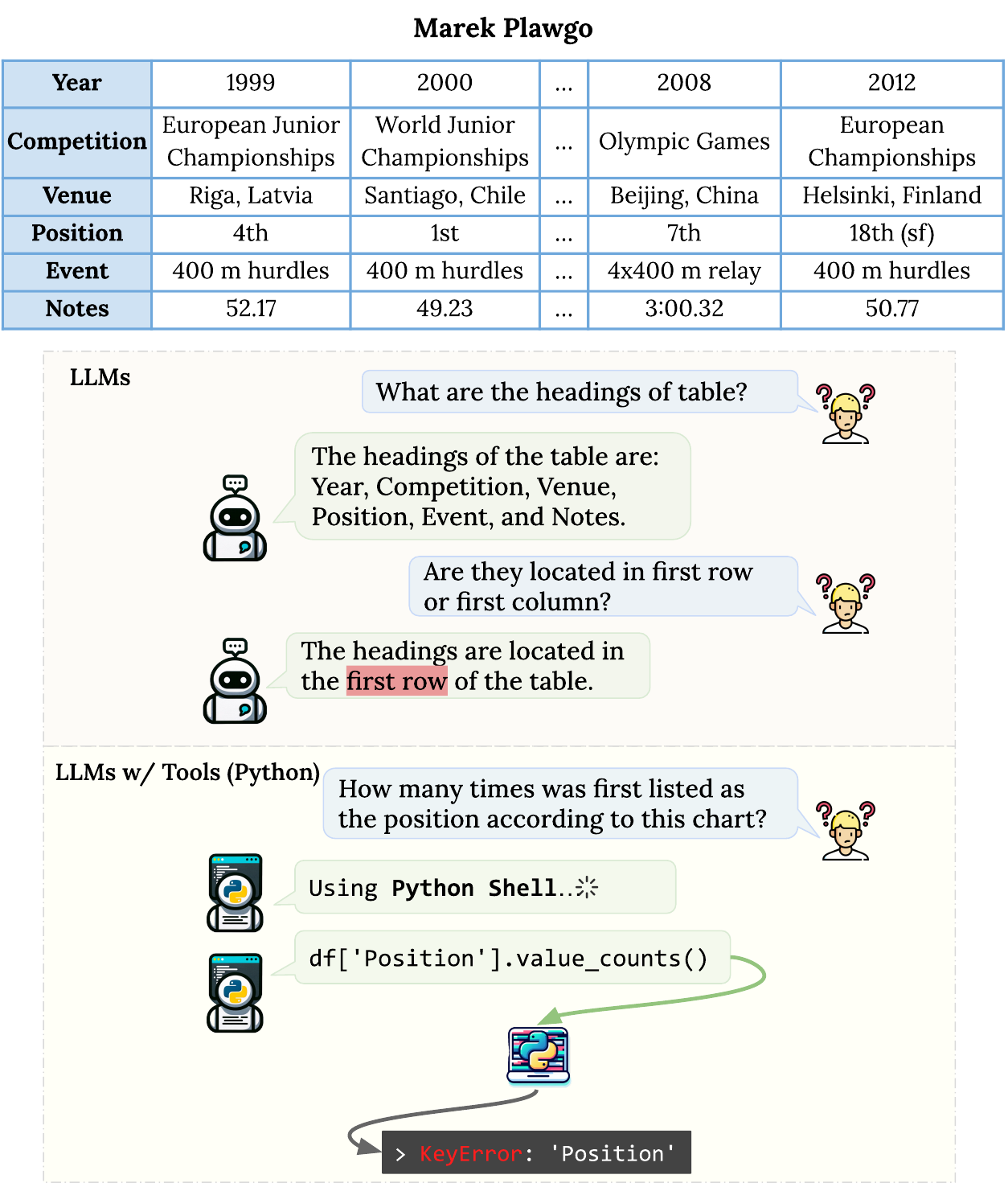}
    \caption{A demonstration of the challenges faced by LLMs in comprehending and interpreting table structures. In \colorbox[HTML]{faede3}{\strut the first example}, despite the LLM correctly identifying table headings, it falters in accurately determining the headings’ positions within the table structure. In \colorbox[HTML]{fafae3}{\strut the second example}, the model using Python Shell as an external tool erroneously defaults to interpreting headings (located in first column) as column headers, leading to subsequent mistakes in the generated code. Some logos in this and subsequent figures are generated by OpenAI's DALL-E3~\cite{openai2023dalle3}.}
    \label{fig:intro_small}
    \vspace{-1em}
\end{figure}

\begin{figure*}[!h]
    \centering
    \vspace{-0.6em}
    \includegraphics[width=\textwidth]{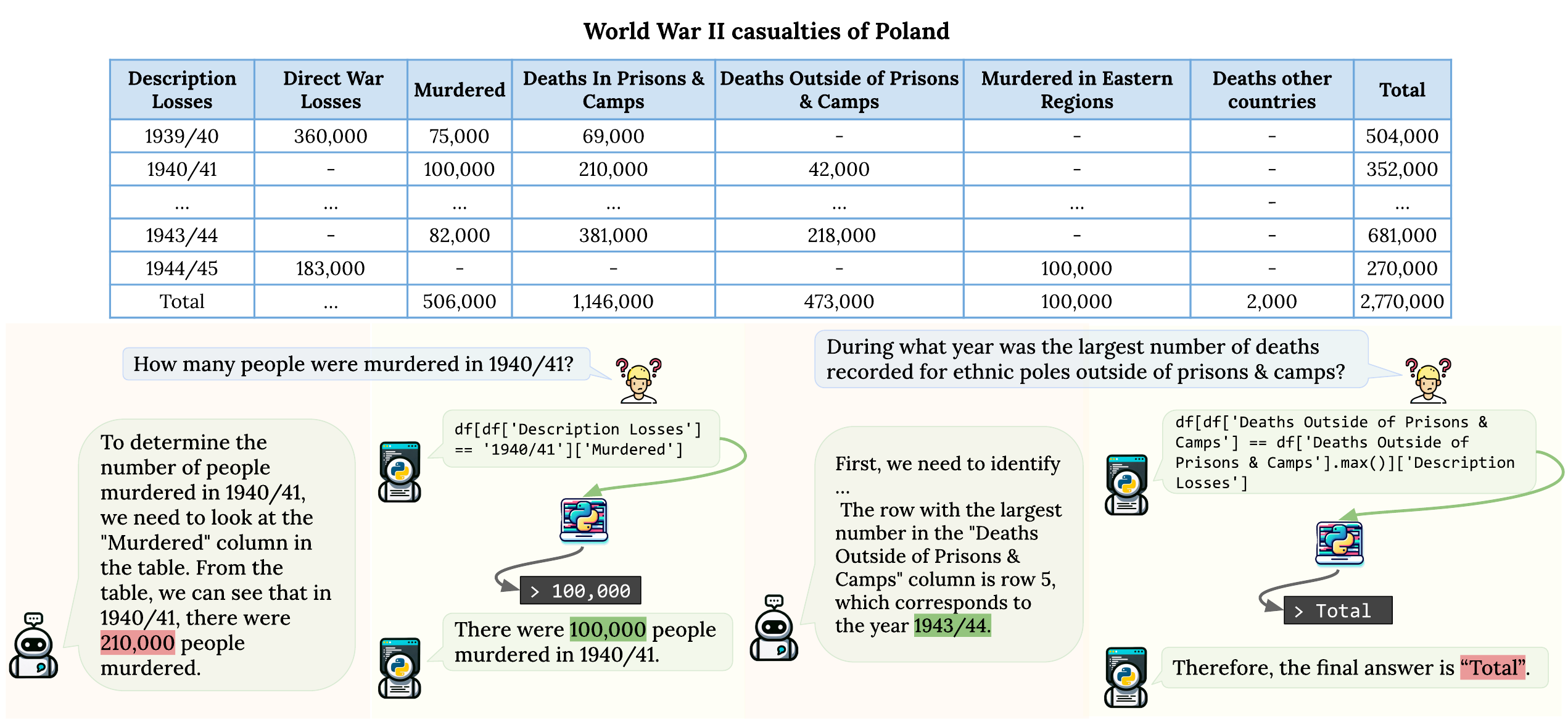}
    \caption{Illustrative examples sampled from the \textsc{WikiTableQuestions} dataset, wherein a comparison is exhibited between \colorbox[HTML]{faede3}{\strut textual reasoning via direct prompting} and \colorbox[HTML]{fafae3}{\strut symbolic reasoning through Python Shell interactions}. \textit{\textbf{Top}}: The table and its title. \textit{\textbf{Bottom Left}}: The first question example where \colorbox[HTML]{faede3}{\strut textual reasoning} erroneously interprets due to the limitation of precision localization, while \colorbox[HTML]{fafae3}{\strut symbolic reasoning} accurately locates the answer with Python code. \textit{\textbf{Bottom Right}}: The second question example where \colorbox[HTML]{faede3}{\strut textual reasoning} successfully identifies the answer, but \colorbox[HTML]{fafae3}{\strut symbolic reasoning} mistakenly treats the special row \texttt{total} row as the final answer.}
    \label{fig:intro_large}
     \vspace{-1.3em}
\end{figure*}

Like humans, LLMs can also benefit from reading tabular data accompanying text. However, as indicated in \Cref{fig:intro_small}, the structural nature of tables presents unique challenges to these models. Inherently designed to parse and process vast expanses of unstructured textual content, LLMs confront a paradigm shift when facing tabular data. Linearizing tables to suit the LLM paradigm can obscure the inherent structural and relational information, making tasks such as precise localization and complex statistical analyses. Additionally, the design variations in tables, whether `column tables' with headers in the first row or `row tables' with headers in the first column, further complicate the interpretation process. Beyond  structural concerns, numerical reasoning and aggregation over tabular data present another layer of complexity. 
While LLMs excel at textual understanding, they occasionally stumble when confronted with tasks necessitating precise numerical computation within tables. Moreover, tables often present a dense amalgamation of textual or numerical data. The sheer volume and intricacy of this information can risk overshadowing crucial details, potentially impeding the LLM's decision-making abilities~\cite{shi2023large}.

With the emergence of instruction fine-tuning techniques \cite{wei2022finetuned, chung2022scaling} and the application of Reinforcement Learning from Human Feedback (RLHF) \cite{stiennon2022learning, gao2022scaling, paul2017deep}, LLMs have witnessed significant enhancements in their alignment capabilities, paving the way for transitioning from few-shot to zero-shot learning settings \cite{kojima2023large}. In light of these advancements, this paper delves deep into the the challenges and intricacies of tabular understanding and reasoning by LLMs, exemplified in \Cref{fig:intro_large}. We organize our exploration around three \textbf{pivotal research questions}: (1) How well do LLMs perceive table structures and how can we ensure robustness against structural variations? (2) Comparing textual and symbolic reasoning for table data in LLMs, which prevails in effectiveness, and what advantages and challenges manifest in each strategy? (3) Will the aggregation of multiple reasoning pathways enhance the accuracy and reliability of tabular data interpretation by LLMs?

In pursuit of answering the aforementioned research questions, we conduct experiments on SOTA LLMs such as GPT-3.5~\cite{openai2023chatgpt}. Our findings in \Cref{rq1} underscore that while LLMs are adept at semantically interpreting tables, their capability to resist structural variance (\Cref{impact_of_table_perturbation}) and understand table structures (\Cref{limitations_of_table_transposition}) is suboptimal.  Motivated by these findings, we propose a table structure normalization method to enhance LLMs’ resilience against structural table variations in \Cref{norm}. Intriguingly, \Cref{rq2_results} reveals that textual reasoning surpasses symbolic reasoning in contexts with limited table content, defying conventional conceptions of symbolic reasoning's dominance in other domains~\cite{mialon2023augmented}. Both textual and symbolic reasoning strategies encompass different advantages and challenges, which is detailed in \Cref{rq2_error}. To harness the unique strengths of each, we implement mix self-consistency mechanism (\Cref{rq3}) that remarkably attains SOTA performance on Table QA, exemplifying the synergistic potential when both reasoning strategies are aggregated.

\section{Related Work}

\stitle{PLMs for Tabular Data Processing.}
Tabular reasoning presents unique challenges due to the fusion of free-form natural language questions with structured or semi-structured tabular data, for which PLMs jointly trained on tables and text are developed in the past few years, including TaBERT \cite{yin-etal-2020-tabert}, TaPas \cite{tapas}, TAPEX \cite{liu2022tapex}, ReasTAP \cite{zhao-etal-2022-reastap}, and PASTA \cite{gu-etal-2022-pasta}.
Despite these advancements, recent studies have identified generalization issues under table perturbations~\cite{zhao2023robut, chang2023drspider}, raising concerns regarding the robustness of PLMs.
Specific efforts like LETA~\cite{zhao2023robut} and LATTICE~\cite{wang2022robust} have investigated and mitigated the vulnerabilities related to structural perturbations of tabular data, like row/column shuffling and table transpose, through various techniques, including data augmentation and order-invariant graph attention. However, these approaches require whitebox access to the models, limiting their applicability to SOTA LLMs with only blackbox accessibility, a limitation directly addressed in this work.

\stitle{Tabular Data Processing with LLMs.}
Recent advancements in LLMs, notably within few-shot learning, have demonstrated their potential for tabular reasoning. \citet{chen2023large} leveraged the Chain-of-Thought (CoT) technique~\cite{wei2023chainofthought} to illustrate LLMs’ effectiveness in this domain. Building upon CoT, \citet{cheng2023binding} and \citet{dater} introduced frameworks that incorporate symbolic reasoning for improved comprehension, with \citeauthor{dater} emphasizing their ability to adeptly decompose both evidence and questions. The advent of aligned models, such as ChatGPT, has enabled zero-shot table reasoning. However, these models often lack sensitivity to table structures, struggling with structural perturbations. StructGPT~\cite{Jiang-StructGPT-2022}, while introducing a promising framework for LLMs to efficiently engage with structured data, has its effectiveness limited by not integrating symbolic reasoning, a critical aspect for enhancing the full capabilities of LLMs in tabular reasoning, which is the focal point of this study. Furthermore, while programming-based approaches can mitigate some challenges, they are limited in addressing free-form queries, creating a gap in the landscape. Innovations like AutoGPT~\cite{Significant_Gravitas_Auto_GPT} have sought to address this, spawning the development of tabular agents like LangChain~\cite{Chase_LangChain_2022}, SheetCopilot~\cite{li2023sheetcopilot}, and DataCopilot~\cite{zhang2023datacopilot}. These agents offer solutions unattainable through conventional programming but still require rigorous evaluation in various scenarios. In our study, we delve into addressing these challenges for enhancing LLMs' reasoning capabilities within structural perturbations, hence providing insights that facilitate improved accuracy in the current context.

\section{Preliminaries}

This section succinctly introduces the foundational aspects of our study over structurally perturbed tabular data. \Cref{PD} formally defines the problem, delineating the critical notations and conceptual frameworks, and \Cref{setup} explicates our experimental setup details, elucidating dataset choice, model utilization, and evaluation strategy.


\subsection{Problem Definition}
\label{PD}

Question answering (QA) over tabular data, commonly known as the TableQA task, is an important challenge in NLP. In this study, we targets TableQA to explore and enhance the proficiency of LLMs, in reasoning over tabular data. Additionally, we probe the robustness and adaptability of these models by introducing structural perturbations to tables.

Let $\mathcal{T}$ represent a table consisting of $\mathcal{R}$ rows and $\mathcal{C}$ columns, and $\tau$ represent its title/caption. Each cell in $\mathcal{T}$ is denoted by $\mathcal{T}_{i,j}$, where $i \in [0, \mathcal{R}-1]$ and $j \in [0, \mathcal{C}-1]$. $\mathcal{T}_{0,j}$ are headers. Given a question $\mathcal{Q}$ pertaining to the table, our task is to identify an answer $\mathcal{A}$. This answer is generally a collection of values, denoted as $\{a_1, a_2, \dots, a_k\}$, where $k \in \mathbb{N}^+$.

Furthermore, to delve deeper into the structural comprehension of LLMs, we introduce structural perturbations, which include:\footnote{Column shuffling was not employed as the typical number of columns is limited and this shuffling had minimal impact on accuracy~\cite{zhao2023robut}.}

\begin{enumerate}[leftmargin=1em]
    \setlength\itemsep{0em}
    \item \textbf{Transposed Table ($\mathcal{T}^{\top}$):} A table obtained by converting rows to columns and vice-versa, maintaining the row and column order:
    \[
    \setlength{\arraycolsep}{0pt} 
    \mathcal{T}^{\top}_{i,j} = \mathcal{T}_{j,i} \quad \forall i \in [0, \mathcal{R}-1], j \in [0, \mathcal{C}-1].
    \]
    
    \item \textbf{Row Shuffled Table ($\mathcal{T}_\Pi$):} A table obtained by randomly shuffling the rows (excluding the headers) with a random permutation function $\pi$, while keeping the order of columns unchanged:
    \[
    \mathcal{T}_{\Pi_{i,j}} = \mathcal{T}_{\pi(i),j} \quad \forall i \in [1, \mathcal{R}-1], j \in [0, \mathcal{C}-1]
    \]
    
    \item \textbf{Row Shuffled and Transposed Table ($\mathcal{T}_{\Pi}^{\top}$):} A table obtained by first randomly shuffling the rows (excluding headers) and then applying transposition:
    \[
    \mathcal{T}^{\top}_{\Pi_{i,j}} = \mathcal{T}_{j,\pi(i)} \quad \forall i \in [1, \mathcal{R}-1], j \in [0, \mathcal{C}-1]
    \]

\end{enumerate}

Defining our research problem more formally: our primary objective is to investigate the function, \( f \), that can appropriately answer the posed question using the provided table. Specifically, this function will take three arguments: the table variant $\mathcal{T'} \in \{\mathcal{T}, \mathcal{T}^{\top}, \mathcal{T}_\Pi,  \mathcal{T}_\Pi^{\top} \}$, its title $\tau$, and the question $\mathcal{Q}$. It will output an answer $\mathcal{A}$. The entire problem can be formally framed as:
\[
f(\mathcal{T'}, \tau, \mathcal{Q}) \rightarrow \mathcal{A}, \quad \forall \mathcal{T'} \in \{\mathcal{T}, \mathcal{T}^{\top}, \mathcal{T}_\Pi,  \mathcal{T}_\Pi^{\top} \}
\]

\subsection{Experimental Setup}
\label{setup}

This section details the experimental setup adopted in our study, including the datasets employed, model selection, evaluation metrics, reasoning methods, and other details.

\stitle{Dataset.} We used the \textsc{WikiTableQuestions} (\textsc{WTQ};~\citealt{pasupat2015compositional}) dataset for our experiments. The test set comprises 421 tables. Each table provides up to two question-answer pairs; if a table has fewer than two, only one was chosen, totaling 837 unique data points. With our four table configurations (original and three perturbations), the overall evaluation data points amount to \( 837 \times 4 = 3,348 \).

\stitle{Models.} We employ the GPT-3.5~\cite{openai2023chatgpt} series for our research. Given that tables usually have extensive data, 
depending on the prompt length, we dynamically use \texttt{gpt-3.5-turbo-0613} and \texttt{gpt-3.5-turbo-16k-0613}, with a primary aim to optimize cost when querying the API.

\stitle{Evaluation Metrics.} 
Following prior works~\cite{jiang-etal-2022-omnitab, ni2023lever, Binder, dater}, we employ \textit{Exact Match Accuracy} as the evaluation metric to validate predictions against ground truths, embedding instructions in prompts for consistent and parseable outputs.

\stitle{Reasoning Methods.}
Our evaluation hinges on two distinct zero-shot reasoning approaches:
\begin{itemize}[leftmargin=1em]
    \setlength\itemsep{0em}
    \item \textbf{Direct Prompting (DP)} is a textual reasoning method that prompts LLMs to answer questions in a zero-shot manner. Rather than directly providing the answer, LLMs are instructed to reason step-by-step before concluding. More details can be found in \Cref{prompt_dp}, 

    \item \textbf{Python Shell Agent (PyAgent)} is a symbolic reasoning approach where the model dynamically interacts with a Python shell. Specifically, LLMs use the Python Shell as an external tool to execute commands, process data, and scrutinize results, particularly within a pandas dataframe, limited to a maximum of five iterative steps. Detailed prompt is presented in \Cref{prompt_agent}.
\end{itemize}

\stitle{Other Details.} Depending on the scenario, we adjust the temperature setting. In cases not employing self-consistency, we set it to 0. For scenarios involving self-consistency, the temperature is set to 0.8. For further granularity, \Cref{appendix/prompts} offers an exhaustive list of the prompts implemented in our experiments. Importantly, it should be noted that all prompts are deployed in a zero-shot manner, without any demonstrations or examples.
\section{LLM Robustness to Structural Perturbations}
\label{rq1}

This section explores how LLMs interpret varied table structures in response to our first research question (\Cref{sec:intro}). We probe the impact of three table perturbations on LLM performance (\Cref{impact_of_table_perturbation}), uncover LLMs’ challenges and limitations for direct table transposition and recoganize tranposed tables (\Cref{limitations_of_table_transposition}), and introduce a structure normalization strategy (\textsc{Norm}) to mitigate these issues (\Cref{norm}).

\subsection{Impacts of Table Perturbations on LLMs}
\label{impact_of_table_perturbation}

\begin{table}[t]
\centering
\small
\renewcommand{\arraystretch}{1.1}
\begin{tabular}{lcc}
\toprule
\textbf{Perturbation} & \textbf{DP} & \textbf{PyAgent} \\
\midrule
Original \( (\mathcal{T}) \) & 59.50 & 55.91 \\
+Shuffle \( (\mathcal{T}_{\Pi}) \) & \cellcolor{rose10}\makecell{52.21 \\ {\scriptsize \textcolor{darkred}{-12.25\%}}} & \cellcolor{rose10}\makecell{47.91 \\ {\scriptsize \textcolor{darkred}{-14.31\%}}} \\
+Transpose \( (\mathcal{T}^{\top}) \) & \cellcolor{rose10}\makecell{51.14 \\ {\scriptsize\textcolor{darkred}{-14.05\%}}} & \cellcolor{rose30}\makecell{12.45 \\ {\scriptsize\textcolor{darkred}{-77.73\%}}} \\
+Transpose\&Shuffle \( (\mathcal{T}^{\top}_{\Pi}) \) & \cellcolor{rose20}\makecell{37.51 \\ {\scriptsize\textcolor{darkred}{-36.96\%}}} & \cellcolor{rose45}\makecell{8.96 \\ {\scriptsize\textcolor{darkred}{-83.97\%}}} \\
\bottomrule
\end{tabular}
\caption{Accuracy of GPT-3.5 under different table perturbations using Direct Prompting (DP) and Python Shell Agent (PyAgent).}
\label{tab:table_perturbations}
\end{table}

In \Cref{PD}, we present three types of structural table perturbations: transposition ($\mathcal{T}^{\top}$), row-shuffling ($\mathcal{T}{\Pi}$), and their combination ($\mathcal{T}^{\top}{\Pi}$). As demonstrated in \Cref{tab:table_perturbations}, both reasoning methods, DP and PyAgent, exhibit significant performance declines, with more pronounced when transposition is applied. DP consistently outperforms PyAgent largely across perturbations, indicating that textual reasoning tends to be more resilient to these structural changes. This resilience can be attributed to LLMs' ability to grasp semantic connections and meanings irrespective of structural shifts. In contrast, symbolic reasoning, exemplified by PyAgent, is heavily reliant on table structure, making it more vulnerable, especially to transposition.

\begin{table}[t]
    \small
    \centering
    \begin{tabular}{llc}
    \toprule
    \textbf{LLMs As} & \textbf{Task Description} & \textbf{Accuracy}\\
    \midrule
    \multirow{2}{*}{Transposer} & \(f(\mathcal{T}) \to \mathcal{T}^{\top}\) & 53.68 \\
    & \(f(\mathcal{T}^{\top}) \to \mathcal{T}\) & 51.07 \\
    \midrule
    \multirow{2}{*}{Detector} & \(f(\mathcal{T}) \to 0\) & 93.35 \\
    & \(f(\mathcal{T}^{\top}) \to 1\) & 32.54 \\
    \midrule
    \multirow{2}{*}{Determinator} & \(f(\mathcal{T}, \mathcal{T}_{0,\ast}, \mathcal{T}_{\ast,0}) \to \mathcal{T}_{0,\ast}\) & 97.39 \\
    & \(f(\mathcal{T}^\top, \mathcal{T}_{0,\ast}, \mathcal{T}_{\ast,0}) \to \mathcal{T}_{\ast,0}\) & 94.77 \\
    \bottomrule
    \end{tabular}
    \caption{Evaluation results of GPT-3.5 on the 421 distinct tables of \textsc{WTQ} dataset, including three tasks: \textit{Transposer} involving switching between original (\(\mathcal{T}\)) and transposed tables (\(\mathcal{T}^{\top}\)), \textit{Detector} for identifying need for table transposition (0 for no transposition, 1 for transposition required), and \textit{Determinator} to choose probable table headings either from the first row (\(\mathcal{T}_{0,\ast}\)) or the first column (\(\mathcal{T}_{\ast,0}\)).}
    \label{tab:combined_results}
\end{table}

\subsection{Limitations of Table Transposition with LLMs}
\label{limitations_of_table_transposition}

To better understand LLMs' capabilities with regards to table structures, we investigate their ability on detecting tables in need of transposition and performing table transposition.

\stitle{LLMs as Transposition Detectors.} Given a table \( \mathcal{T} \), the goal is to detect whether a table should be transposed for better comprehension by LLMs. This is formulated as a binary classification task:
$$
f(\mathcal{T}) \to 0, \quad f(\mathcal{T}^{\top}) \to 1,
$$
Where 0 denotes `no need of transposition' and 1 indicates `transposition needed'.
\Cref{tab:combined_results} shows the results using the prompt in \Cref{prompt_recommender}. GPT-3.5 correctly classified 93.35\% of original tables \( \mathcal{T} \) as not requiring transposition. However, its accuracy dramatically decreased to 32.54\% on transposed tables \( \mathcal{T}^{\top} \). Our observations highlight 
that LLMs suffer from structural bias in the interpretation of table orientations, predominantly leading to recommendations against transposition.

\stitle{LLMs as Table Transposers.} The objective is to switch between original and transposed table formats. Specifically, the goal is to directly yield \( \mathcal{T}^{\top} \) given \( \mathcal{T} \), and vice versa. 
Formally, the task is:
\[
f(\mathcal{T}) \to \mathcal{T}^{\top}, \quad f(\mathcal{T}^{\top}) \to {\mathcal{T}}
\]

We observed that GPT-3.5's proficiency in this task is limited, with an accuracy of 53.68\% transposing row tables and 51.07\% for the inverse operation, suggesting that LLMs can not transpose tables precisely. For a detailed error case study and further analysis, refer to the \Cref{analysis_transposer}.

\subsection{Table Structure Normalization}
\label{norm}

In addressing structural variations in tables, our goal is to ensure consistent interpretation and utility across diverse table structures. To normalize various table structures into well-ordered row-tables prior to downstream tasks, we introduce \textsc{Norm}, which is a two-stage normalization strategy: the first stage detects column-tables and transposing them into row-tables, while the second stage sorts the row-tables for enhanced comprehensibility. Through this approach, \textsc{Norm} accommodates for structural perturbations without compromising the understanding of the standardized row-tables.


\stitle{Content-Aware Transposition Determination} 
In the straightforward methods mentioned in \Cref{limitations_of_table_transposition}, LLMs are affected by the loss of structure information of the table.
Our approach aims to reduce this structural dependence by introducing a content-aware determination process, which leverages the semantic reasoning capabilities of LLMs, instead of perceiving the table's structure. Specifically, we analyze the inherent content within the first row (\(\mathcal{T}_{0,\ast}\)) and the first column (\(\mathcal{T}_{\ast,0}\)) of a given table (\( \mathcal{T} \)) to decide which is more semantically fitting to serve as the table's heading. This content-aware approach can be mathematically modeled as:
$$
\begin{cases} 
f(\mathcal{T}, \mathcal{T}_{0,\ast}, \mathcal{T}_{\ast,0}) \to \mathcal{T}_{0,\ast} \\
f(\mathcal{T}^\top, \mathcal{T}_{0,\ast}, \mathcal{T}_{\ast,0}) \to \mathcal{T}_{\ast,0} \\
\end{cases}
$$
Here, a selection of the first row suggests that the current table structure is preferred, whereas opting for the first column signifies a need for transposition.
The prompt detailing this method is provided in \Cref{prompt_transpose_check}. Results in \Cref{tab:combined_results} highlight capability of GPT-3.5 in discerning table headings semantically, with accuracies of 97.39\% and 94.77\% respectively for original table and tranposed table.

\begin{table}[!t]
\small
\centering
\begin{tabular}{lcccc}
\toprule
\textbf{Method} & \textbf{\( \mathcal{T} \)} & \textbf{\(\mathcal{T}_{\Pi}\)} & \textbf{\( \mathcal{T}^{\top}\)} & \textbf{\( \mathcal{T}_{\Pi}^{\top}\)} \\
\midrule
DP & 59.50 & 52.21 & 51.14 & 37.51 \\
\quad\multirow{2}{*}{+\textsc{Norm}} & \cellcolor{rose5}58.66 & \cellcolor{mint20}58.66 & \cellcolor{mint30}58.30 & \cellcolor{mint50}57.71 \\
& \cellcolor{rose5}{\scriptsize \textcolor{darkred}{-1.41\%}} & \cellcolor{mint20}{\scriptsize \textcolor{darkgreen}{+12.35\%}} & \cellcolor{mint30}{\scriptsize \textcolor{darkgreen}{+14.00\%}} & \cellcolor{mint50}{\scriptsize \textcolor{darkgreen}{+53.85\%}} \\
\midrule
PyAgent & 55.91 & 47.91 & 12.43 & 8.96 \\
\quad\multirow{2}{*}{+\textsc{Norm}} & \cellcolor{mint5}56.87 & \cellcolor{mint20}57.11 & \cellcolor{mint80}55.44 & \cellcolor{mint90}55.08 \\
& \cellcolor{mint5}{\scriptsize \textcolor{darkgreen}{+1.72\%}} & \cellcolor{mint20}{\scriptsize \textcolor{darkgreen}{+19.20\%}} & \cellcolor{mint80}{\scriptsize \textcolor{darkgreen}{+346.02\%}} & \cellcolor{mint90}{\scriptsize \textcolor{darkgreen}{+514.73\%}} \\
\bottomrule
\end{tabular}
\caption{Accuracy of GPT-3.5 under different table perturbations for Direct Prompting (DP) and Python Shell Agent (PyAgent) with \textsc{Norm} applied.}
\label{tab:results_norm}
\end{table}

\stitle{Row Reordering.} Upon transposition, our next objective is to ensure the logical coherence of the table data through reordering the rows. We instruct LLMs to suggest improved reordering strategies using the prompts as detailed in \Cref{prompt_sort}. 
Due to the subjective nature involved in identifying the most suitable order of a tabular data, and given that there are no widely recognized standards for this process, the effectiveness of the proposed sorting strategy will be evaluated based its downstream impact on the results of table QA task.
We notice that when the entire well-ordered table is exposed, GPT-3.5 occasionally suggests alternative sorting strategies, leading to unnecessary complexity. To counteract this tendency and ensure a better sorting proposal, we strategically present the model with only the first three and the last three rows of the table. This selective exposure typically allows the model to discern logical ordering patterns without being influenced by existing table configurations.

\Cref{tab:results_norm} underscores the efficacy of \textsc{Norm} when applied prior to the two reasoning methods -- DP and PyAgent. Demonstrably, \textsc{Norm} robustly mitigates structural perturbations, optimizing table comprehensibility for LLMs.  The results illustrate that applying \textsc{Norm} does not detrimentally affect the original results (\( \mathcal{T} \)), and it effectively refines perturbated data, aligning the outcomes closely with the original results, and in some instances, even showing slight improvement. This suggests that \textsc{Norm} as a preprocessing step for preparing tabular data can enhance robust analysis by LLMs.


In addressing our initial research question, the analysis indicates that \textbf{LLMs’ performance is sensitive to table structural variations}, with significant struggles observed in accurately interpreting 
the same tabular content under transposition and shuffling.
While \textbf{textual reasoning demonstrates some resilience} to structural variations, \textbf{symbolic reasoning is significantly impacted}, particularly with transposed tables. \textbf{The \textsc{Norm} strategy effectively navigates these challenges} by eliminating dependency on table structures, providing consistent interpretation across diverse table structures without compromising the integrity or meaning of the original content.

\section{Comparing Textual and Symbolic Reasoning}
\label{rq2}

In this section, we delve into the comparison of textual and symbolic reasoning methods in LLMs for tabular data understanding (\Cref{rq2_results}), further conducting a detailed error analysis (\Cref{rq2_error}) to address the second research question (\Cref{sec:intro}). We evaluate the performance of each reasoning strategy using GPT-3.5, shedding light on their strengths and challenges. In \Cref{norm}, we explored \textsc{Norm} to mitigate structural perturbations, enhancing generalized LLM performance and successfully restoring perturbed tables to accuracy levels similar to their original states. Therefore, subsequent analyses will exclusively consider the original tables (\(\mathcal{T}\)).

\subsection{Results}
\label{rq2_results}

\begin{table}[t]
    \centering
    \small
    \begin{tabular}{lc}
    \toprule
    \textbf{Method} & \textbf{Accuracy (\%)} \\
    \midrule
    \rowcolor{mygray} 
    \multicolumn{2}{c}{\centering\textit{Few-shot Prompting Methods}} \\
    \textsc{Binder}$^\bigstar$~\cite{Binder} & 63.61 \\
    \textsc{Binder}$^\spadesuit$~\cite{Binder} & 55.07\\
    \textsc{DATER \smallbold{w/o SC}}$^\bigstar$~\cite{dater} & 61.75 \\
    \textsc{DATER \smallbold{w/ SC}}$^\bigstar$~\cite{dater} & 68.99 \\
    \midrule
    \rowcolor{mygray} 
    \multicolumn{2}{c}{\centering\textit{Zero-shot Prompting Methods}} \\
    \textsc{StructGPT}$^\spadesuit$~\cite{Jiang-StructGPT-2022} & 51.77 \\
    \noalign{\vskip 1.0pt}
    \hdashline
    \noalign{\vskip 1.0pt}
    \textsc{Norm+DP}$^\spadesuit$ & 58.66 \\
    \textsc{Norm+PyAgent}$^\spadesuit$ & 56.87 \\
    \textsc{Norm+PyAgent-Omitted}$^\spadesuit$ & 52.45 \\
    \textsc{Norm+DP\&PyAgent \smallbold{w/ Eval}}$^\spadesuit$ & 64.22 \\
    \noalign{\vskip 1.0pt}
    \hdashline
    \noalign{\vskip 1.0pt}
    \textsc{DP \smallbold{w/ SC}}$^\spadesuit$ & 66.39 \\
    \quad\textsc{+Norm}$^\spadesuit$ & 64.10 \\
    \quad\textsc{+Norm \smallbold{w/o Resort}}$^\spadesuit$ & 66.99 \\
    \noalign{\vskip 1.0pt}
    \hdashline
    \noalign{\vskip 1.0pt}
    \textsc{PyAgent \smallbold{w/ SC}}$^\spadesuit$ & 61.39 \\
    \quad\textsc{+Norm}$^\spadesuit$ & 63.77 \\
    \quad\textsc{+Norm \smallbold{w/o Resort}}$^\spadesuit$ & 62.84 \\
    \noalign{\vskip 1.0pt}
    \hdashline
    \noalign{\vskip 1.0pt}
    \textsc{DP\&PyAgent \smallbold{w/ Mix-SC}}$^\spadesuit$ & 73.06 \\
    \quad\textsc{+Norm}$^\spadesuit$ & 72.40 \\
    \quad\textsc{+Norm \smallbold{w/o Resort}}$^\spadesuit$ & 73.65 \\
    \bottomrule
    \end{tabular}
    \caption{Performance on the sampled \textsc{WTQ} dataset ($\mathcal{T}$). $^\bigstar$ denotes methods based on Codex, while $^\spadesuit$ represents those based on GPT-3.5. The term SC refers to self-consistency.\tablefootnote{For SC, results are derived by conducting an average over 100 shuffles to accommodate instances of ties during majority voting. In the Mix-SC method, DP-derived answers are prioritized over PyAgent due to DP's observed superior performance. All experiments regarding SC follow this.} \textsc{Norm \smallbold{w/o Resort}} means that reordering stage for \textsc{Norm} is not performed.}
    \label{tab:all_results}
\end{table}

\begin{table*}[t]
\centering
\small
\setlength\tabcolsep{4pt}
\begin{tabular}{lccm{6.5cm}m{3.6cm}}
\toprule
\textbf{Error Types} & \textbf{DP} & \textbf{PyAgent} & \textbf{Description} & \textbf{Case Study} \\
\midrule
Table Misinterpretation & 42\% & -$^\dagger$ & LLMs incorrectly interpret the content in tables. & \Cref{counting_error}, \Cref{locating_error}\\
\midrule
Coding Errors & - & 38\% & LLMs produce inaccurate code, typically due to issues with minor details. & \Cref{noise_error}, \Cref{special_row_error}, \Cref{wrong_coding} \\
\midrule
Misalignment Issue& 24\% & 28\% & Outputs are conceptually correct but the answers do not align with the instructions. & \Cref{format_error}, \Cref{deviation_error} \\
\midrule
Logical Inconsistency & 20\% & 10\% & LLMs exhibit failures in reasoning, leading to contradictions or inconsistencies. & \Cref{reasoning_error_dp}, \Cref{reasoning_error_pyagent} \\
\midrule
Execution Issue & - & 12\% & Issues emerge related to the execution of Python code. & \Cref{loop_error}, \Cref{non_observable_error}\\
\midrule
Resorting Issue & 10\% & 8\% & The resorting stage in \textsc{Norm} changes the answers of some sequence-dependent questions. & \Cref{norm_error} \\
\bottomrule
\end{tabular}
\caption{Error types of DP and PyAgent methods. $^\dagger$This does not imply that PyAgent does not make \textit{table interpretation} errors; these are included under \textit{coding errors} to avoid overlapping. Note that the percentages for each reasoning method might not sum up to 100\%; the remaining percentage points are attributable to \textit{other errors}, such as problems with dataset labeling, which are not categorized here.}
\label{tab:error_analysis}
\end{table*}

\Cref{tab:all_results} showcases the performance of GPT-3.5 when employed for both direct textual reasoning using DP and interactive symbolic reasoning using PyAgent. By instructing the model with the CoT~\cite{wei2023chainofthought} reasoning strategy to \textit{think step by step, and then give the final answer}, as detailed in \Cref{prompt_dp}, we can achieve an accuracy of 58.66\%. This surpasses the StructGPT's Iterative Reading-then-Reasoning method, which concentrates reasoning tasks by continually collecting relevant evidence. For tables with limited tokens, symbolic reasoning via PyAgent offers an accuracy of 56.87\%, which is slightly behind the accuracy by DP in a single attempt. A distinct advantage of symbolic reasoning is its ability to only present parts of the table in the prompt. As our experiments revealed, after excluding the central rows and showcasing only the initial and final three rows, we manage to maintain an accuracy of 52.45\% with a 4.42\% drop compared to the full-table PyAgent results. This makes it possible to deal with larger tables with numerous rows using LLMs with limited context window. In the following sections, we will present a comprehensive analysis of the discrepancies and errors observed across these methods.

\subsection{Error Analysis}
\label{rq2_error}

To elucidate the challenges and limitations of DP and PyAgent, this section presents an in-depth error analysis by sampling 50 erroneous outputs for each. \Cref{tab:error_analysis} summarizes the predominant error types for DP and PyAgent methods. \textit{Table interpretation errors} significantly afflict the DP method, comprising 42\% of its total errors, highlighting substantial challenges for LLMs in accurately interpreting table data. PyAgent primarily struggles with \textit{coding errors}, constituting 38\% of its total errors. These errors either originate from misunderstandings of table content, often overlooking subtle details, or manifest as inherent deficiencies in coding capabilities. These prevalent errors underscore the intrinsic challenges and limitations each method faces in the reasoning process. Detailed case studies on each error type are delineated in \Cref{appendix/error_dp_pyagent}.

In response to the second research question, the analysis indicates \textbf{DP marginally surpasses PyAgent within single attempts}. Despite this, PyAgent can handle larger tables by processing partial table views. Notably, \textbf{DP encounters difficulties in accurate table interpretation}, while \textbf{PyAgent reveals instability in coding capabilities}.

\section{Reasoning Aggregation}
\label{rq3}

This section examines how combining multiple reasoning pathways can boost LLMs' accuracy in interpreting tabular data, which is in response to the third research question (\Cref{sec:intro}).

\subsection{Methods}

\stitle{Self-Consistency.} Previous work has highlighted the advantages of generating multiple outputs from LLMs and adopting the most frequent answer, a mechanism known as self-consistency (SC;~\citealt{wang2023selfconsistency}). \Cref{tab:all_results} showcases the notable improvements realized through self-consistency (aggregating 10 outputs), with DP achieving an accuracy of 64.84\% and PyAgent attaining 63.49\%.

\stitle{Self-Evaluation.}
Based on our error analysis in \Cref{rq2_error}, different reasoning methods excel at specific tasks. For instance, symbolic reasoning tends to outperform textual reasoning in counting and column localization tasks. To optimize the choice between these methods, we strategically use a prompt (referenced in \Cref{prompt_self_eval}), which avoids directly validating answers against tables but guides the LLM to choose between the two reasoning approaches based on the question's nature and each answer's clarity. By weighing the problem against the known strengths and weaknesses of each reasoning strategy, this tactic mitigates potential bias towards textual reasoning by LLMs and enhances answer accuracy. As evidenced by \Cref{tab:all_results}, using self-evaluation boosts accuracy to 64.99\%. Impressively, this method, using only two reasoning paths, matches the performance of using 10 paths of DP or PyAgent independently.

\stitle{Mix Self-Consistency.} According to \Cref{rq2_results}, symbolic and textual reasoning exhibit distinct focuses but deliver similar performance. Consequently, we introduce \textit{Mix Self-Consistency}, a method that selects a predetermined number of outputs for each type of inference, aiming for self-consistency. 
This approach hinges on the idea that multiple outputs can reflect the confidence levels of LLMs in answer generation. In scenarios where LLMs are less proficient, they tend to produce a diverse set of answers. Conversely, for tasks that LLMs handle adeptly, consistent answers are often generated across multiple reasoning attempts, converging towards one answer. Such convergence allows for the aggregation of model outputs that align with areas where LLMs exhibit stronger reasoning capabilities, thereby substantially improving accuracy. The detailed mechanics of how this approach is operationalized within the framework of \textit{Mix Self-Consistency}, including the aggregation and interpretation of these outputs, are further elucidated in \Cref{mechanics_of_msc}.
\Cref{tab:all_results} demonstrates that using mix self-consistency (generating 5 outputs per inference type,\footnote{The choice of generating 5 outputs per inference type (5+5) is a hyperparameter selection influenced by the dataset's distribution. We conducted an ablation study regarding this in~\Cref{ablation_output_selection}. We use an equal split (5+5) based on observed comparable performance between the two reasoning strategies.} totaling 10) enhances performance substantially, achieving an impressive accuracy of 72.40\%, which achieves SOTA performance on the sampled \textsc{WTQ} data.

\subsection{Overall Evaluation}
\label{all_eval}

To evaluate our method thoroughly, we conduct a comprehensive pass of testing using the complete \textsc{WTQ} test set, integrating both \textsc{Norm} and Mix self-consistency mechanisms. Since re-sorting may change the answers of row index-related questions, we perform \textsc{Norm} without resorting in this evaluation.\footnote{Originally, the \textsc{Norm} process included a re-sorting step to counteract the row-shuffling perturbation. However, re-sorting may inadvertently alter answers reliant on the initial sequence, as explored in the error analysis (\Cref{rq2_error}) with a detailed case study in~\Cref{norm_error}.} However, it is noteworthy that re-sorting can be advantageous for questions not dependent on row indexes, particularly when dealing with tables that are initially unorganized or messy.

\begin{table}[t]
    \centering
    \small
    \begin{tabular}{lc}
    \toprule
    \textbf{Method} & \textbf{Accuracy (\%)} \\
    \midrule
    \rowcolor{mygray} 
    \multicolumn{2}{c}{\centering\textit{Fine-tuning Based Models}} \\
    \textsc{TaPaS}~\cite{tapas} & 48.8\\
    \textsc{T5-3B}~\cite{UnifiedSKG} &  49.3\\
    \textsc{TaPaX}~\cite{liu2022tapex} &  57.5\\
    \textsc{ReasTAP}~\cite{zhao-etal-2022-reastap} & 58.7\\
    \textsc{OmniTab}~\cite{jiang-etal-2022-omnitab} & 63.3\\
    \midrule
    \rowcolor{mygray} 
    \multicolumn{2}{c}{\centering\textit{LLMs Based Methods}} \\
    \textsc{StructGPT}$^\bigstar$~\cite{Jiang-StructGPT-2022} & 48.4 \\
    \textsc{Binder}$^\bigstar$~\cite{Binder} & 55.5\\
    \textsc{Binder}$^\spadesuit$~\cite{Binder} & 64.6\\
    \textsc{Lever}$^\spadesuit$~\cite{ni2023lever} & 65.8\\
    \textsc{Dater}$^\spadesuit$~\cite{dater} & 65.9\\
    \midrule
    \textbf{Ours}$^\bigstar$ &  \textbf{73.6}\\
    \bottomrule
    \end{tabular}
    \caption{Comparison of various methods on all test data of \textsc{WTQ}. $^\bigstar$ denotes methods based on the GPT-3.5~\cite{openai2023chatgpt}; $^\spadesuit$ denotes methods based on the Codex~\cite{openai2022codex}.}
    \label{tab:all_result}
\end{table}

As illustrated in \Cref{tab:all_result}, our proposed method exhibits outstanding efficacy with an accuracy of 73.6\%, significantly outperforming existing models to achieve SOTA performance on the complete \textsc{WTQ} test set. Importantly, our approach is conducted in a fully zero-shot manner. For a detailed analysis of how table size impacts method performance, see \Cref{sec:table_size_performance}.

In response to the third research question, our findings reveal that \textbf{reasoning path aggregation significantly enhances LLMs' accuracy in table reasoning tasks}. Notably, \textbf{the \textit{Mix Self-Consistency} method achieves an accuracy of 73.6\% on the \textsc{WTQ} dataset, surpassing the previous SOTA by a considerable margin}. The \textit{Self-Evaluation} strategy also contributes to this remarkable performance by adeptly selecting between reasoning approaches.

\section{Conclusion}
This study delved into the proficiency of LLMs in tabular reasoning. We identify
that LLMs are sensitive to structural variance of tables, yet the application of a normalization strategy, \textsc{NORM}, can stabilize the table structures, thus reinforcing the resistance to structural perturbations. When it comes to comparing reasoning approaches, textual reasoning demonsrate slight superiority over symbolic reasoning, with each strategy possessing its unique advantages. Moreover, integrating multiple reasoning strategy via mix self-consistency is proved beneficial for overall interpretation accuracy, significantly outperforming previous SOTA results on the \textsc{WTQ} dataset. Collectively, these observations pave the way for refining LLMs' approach to tabular understanding and reasoning.

\section*{Limitation}
While this study provides insights into tabular data reasoning with LLMs, it is pertinent to acknowledge its limitations. First, the exclusive utilization of GPT-3.5, due to the budgetary constraints, may limit the generalizability of our findings, as exploration with GPT-4 might offer enhanced outcomes. Second, all table data are sourced from Wikipedia, which may introduce potential data leakage or memorization issues, as certain answers might be implicitly available within the LLMs’ training data, thus potentially biasing results. Lastly, several perturbation-sensitive table-based questions, especially regarding table perturbations like shuffling, may impact the precision of the reported accuracy, as demonstrated answers may change based on the structural modifications of the table.

\bibliography{anthology,custom}
\bibstyle{acl_natbib}

\appendix
\clearpage

\onecolumn  

\begin{center}
    {\Large\textbf{Appendices}}
\end{center}

\section{Prompts}
\label{appendix/prompts}

\subsection{Prompt of Direct Prompting (DP).}
\label{prompt_dp}

\begin{lstlisting}[breaklines=true,breakatwhitespace=true]
You are an advanced AI capable of analyzing and understanding information within tables. Read the table below regarding "{[TITLE]}".

{[TABLE]}

Based on the given table, answer the following question:

{[QUESTION]}

Let's think step by step, and then give the final answer. Ensure the final answer format is only "Final Answer: AnswerName1, AnswerName2..." form, no other form. And ensure the final answer is a number or entity names, as short as possible, without any explanation.
\end{lstlisting}

\subsection{Prompt of Python Agent.}
\label{prompt_agent}
\begin{lstlisting}[breaklines=true,breakatwhitespace=true]
You are working with a pandas dataframe in Python. The name of the dataframe is `df`. Your task is to use `python_repl_ast` to answer the question posed to you.

Tool description:
- `python_repl_ast`: A Python shell. Use this to execute python commands. Input should be a valid python command. When using this tool, sometimes the output is abbreviated - ensure it does not appear abbreviated before using it in your answer.

Guidelines:
- **Aggregated Rows**: Be cautious of rows that aggregate data such as 'total', 'sum', or 'average'. Ensure these rows do not influence your results inappropriately.
- **Data Verification**: Before concluding the final answer, always verify that your observations align with the original table and question.

Strictly follow the given format to respond:

Question: the input question you must answer
Thought: you should always think about what to do to interact with `python_repl_ast`
Action: can **ONLY** be `python_repl_ast`
Action Input: the input code to the action
Observation: the result of the action
... (this Thought/Action/Action Input/Observation can repeat N times)
Thought: after verifying the table, observations, and the question, I am confident in the final answer
Final Answer: the final answer to the original input question (AnswerName1, AnswerName2...)

Notes for final answer:
- Ensure the final answer format is only "Final Answer: AnswerName1, AnswerName2..." form, no other form. 
- Ensure the final answer is a number or entity names, as short as possible, without any explanation.
- Ensure to have a concluding thought that verifies the table, observations and the question before giving the final answer.

You are provided with a table regarding "{[TITLE]}". This is the result of `print(df.to_markdown())`:

{[TABLE]}

**Note**: All cells in the table should be considered as `object` data type, regardless of their appearance.

Begin!
Question: {[QUESTION]}
\end{lstlisting}

\subsection{Prompt of LLMs as Table Transposer}
\label{prompt_transposer}

\begin{lstlisting}[breaklines=true,breakatwhitespace=true]
You are given the following table:

{[TABLE]}

Please transpose this table. Maintain the format I give, with each row beginning with '|' and each cell separated by ' | '. Do not change the content of any cell. Your response should solely consist of the transposed table, without any additional text.
\end{lstlisting}

\subsection{Prompt of LLMs as Table Transposition Detector}
\label{prompt_recommender}
\begin{lstlisting}[breaklines=true,breakatwhitespace=true]
Please examine the provided table:

{[TABLE]}

To enhance readability and facilitate efficient data analysis, it is often suggested that the table headings be horizontally located in the first/topmost row. 

Please evaluate the table with this consideration in mind, and provide your response in the following format:

**Table Headings**: List the headings of the table, separated by commas.
**Table Evaluation**: Identify whether the headings listed are horizontally located in the first/topmost row. If not, describe the position.
**Transpose Recommended**: Indicate if transposing is recommended. Answer with only "YES" or "NO", without any additional explanation.
\end{lstlisting}

\subsection{Prompt of Content-Aware Transposition Determination}
\label{prompt_transpose_check}
\begin{lstlisting}[breaklines=true,breakatwhitespace=true]
You are an advanced AI capable of analyzing and understanding information within tables. Read the table below regarding "{[TITLE]}".

{[TABLE]}

Headings of a table are labels or titles given to rows or columns to provide a brief description of the data they contain.

Based on the given table, the headings of the table are more likely to be:

(A) {[FIRST_ROW]}
(B) {[FIRST_COLUMN]}
(C) None of the above

Directly give your choice. Ensure the format is only "Choice: (A)/(B)/(C)" form, no other form, without any explanation.
\end{lstlisting}

\subsection{Prompt of Resorting}
\label{prompt_sort}
\begin{lstlisting}[breaklines=true,breakatwhitespace=true]
You are an advanced AI capable of analyzing and understanding information within tables. Read the table below regarding "{[TITLE]}":

{[TABLE]}

Note: Only selected rows from the beginning and end of the table are displayed for brevity. Intermediate rows are omitted and represented by "..." for clarity.

The table column headings are provided below, separated by semicolons:

{[HEADINGS]}

In order to optimize the interpretability and readability of the data, follow these guidelines to determine the most suitable sorting method:

Sorting Guidelines:

1. Evaluate columns based on data types such as numerical, alphabetical, chronological, categorical, or other relevant sorting methods.
2. Identify any patterns or relationships in the data that would be highlighted by certain sorting methods.
3. Consider column position, as those on the left may sometimes have sorting priority.
4. If applicable, consider sorting by multiple columns in a prioritized sequence.

Provide your decision using one of the following statements:

- For sorting using a single column: "Sort by: [Name of Column]".
- For sorting using multiple columns: "Sort by: [Primary Column Name], [Secondary Column Name], ...".
- If no specific sorting seems advantageous: "Sort by: N/A".

Your response should strictly follow the formats provided.
\end{lstlisting}

\subsection{Prompt of Self-Evaluation}
\label{prompt_self_eval}
\begin{lstlisting}[breaklines=true,breakatwhitespace=true]
Below is a markdown table regarding "{[TITLE]}":

{[TABLE]}

You're tasked with answering the following question:

{[QUESTION]}

You have 2 answers derived by two different methods. Answer A was derived by prompting the AI to think step-by-step. Answer B was derived by interacting with a Python Shell.

Answer A is {[COT_ANSWER]}.
Answer B is {[AGENT_ANSWER]}.

Your task is to determine which is the correct answer. It is crucial that you strictly adhere to the following evaluation process:

1. **Preliminary Evaluation**: Begin by evaluating which of the two answers directly addresses the question in a straightforward and unambiguous manner. A direct answer provides a clear response that aligns closely with the query without introducing additional or extraneous details. If one of the answers is not a direct response to the question, simply disregard it.
2. **Nature of the Question**: If both answers appear to be direct answers, then evaluate the nature of the question. For tasks involving computation, counting, and column-locating, especially when for extensive table, the Python Shell (Answer B) might be more precise. However, always remain cautious if the Python Shell's output appears off (e.g., error messages, success notifications, etc.). Such outputs may not be trustworthy for a correct answer.
3. **Final Verdict**: Finally, after thorough evaluation and explanation, provide your verdict strictly following the given format:
  - Use "[[A]]" if Answer A is correct.
  - Use "[[B]]" if Answer B is correct.

Note: 
1. Each method has its own strengths and weaknesses. Evaluate them with an unbiased perspective. When in doubt, consider the nature of the question and lean towards the method that is most suited for such queries.
2. Ensure that your verdict is provided after evaluation, at the end.
\end{lstlisting}

\newpage

\section{Analysis for LLMs as Table Transposer}
\label{analysis_transposer}

\subsection{Case Study}

\definecolor{bluecell}{HTML}{005782}
\definecolor{greencell}{HTML}{008000}
\definecolor{purplecell}{HTML}{4B0082}
\definecolor{redhighlight}{HTML}{f8d7da}

\begin{figure}[!h]
    \centering
    \includegraphics[width = \textwidth]{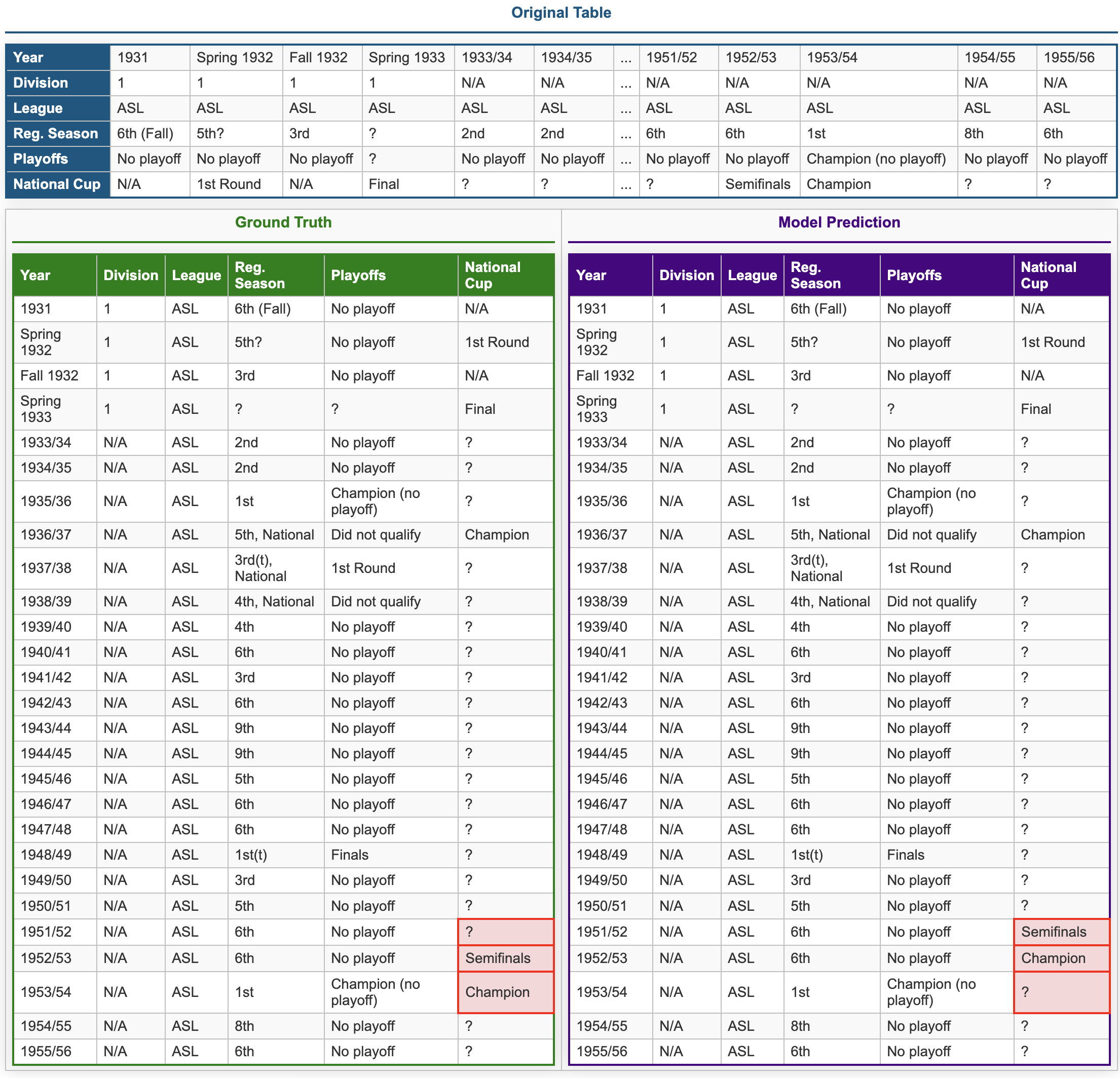}
    \caption{An example error case of content misalignment occurring within cells when leveraging LLMs directly to transpose a table. \textcolor{bluecell}{\textbf{Blue Table (Top):}} The original table subjected to a transposition operation. \textcolor{greencell}{\textbf{Green Table (Buttom Left):}} The ground truth table subsequent to transposition. \textcolor{purplecell}{\textbf{Purple Table (Buttom Right):}} GPT-3.5’s output of transposed table. Cells erroneously aligned or displaced are \colorbox{redhighlight}{highlighted in red}.}
    \label{fig:error_transpose}
\end{figure}

\Cref{fig:error_transpose} illustrates a typical mistake made by LLMs when transposing tables, a problem that becomes more evident when a table has many identical or similar entries. Take, for example, the 'Nation Cup' column shown in the figure, which is filled with numerous \texttt{?} symbols. LLMs, limited in processing structured data, often mishandle such tables, leading to misplacements or misalignments. This highlights the fundamental difficulties and limitations LLMs face in accurately transposing tables containing repetitive or similar data cells.

\subsection{Analysis}

\begin{figure*}[!h]
    \centering
    \includegraphics[width=\textwidth]{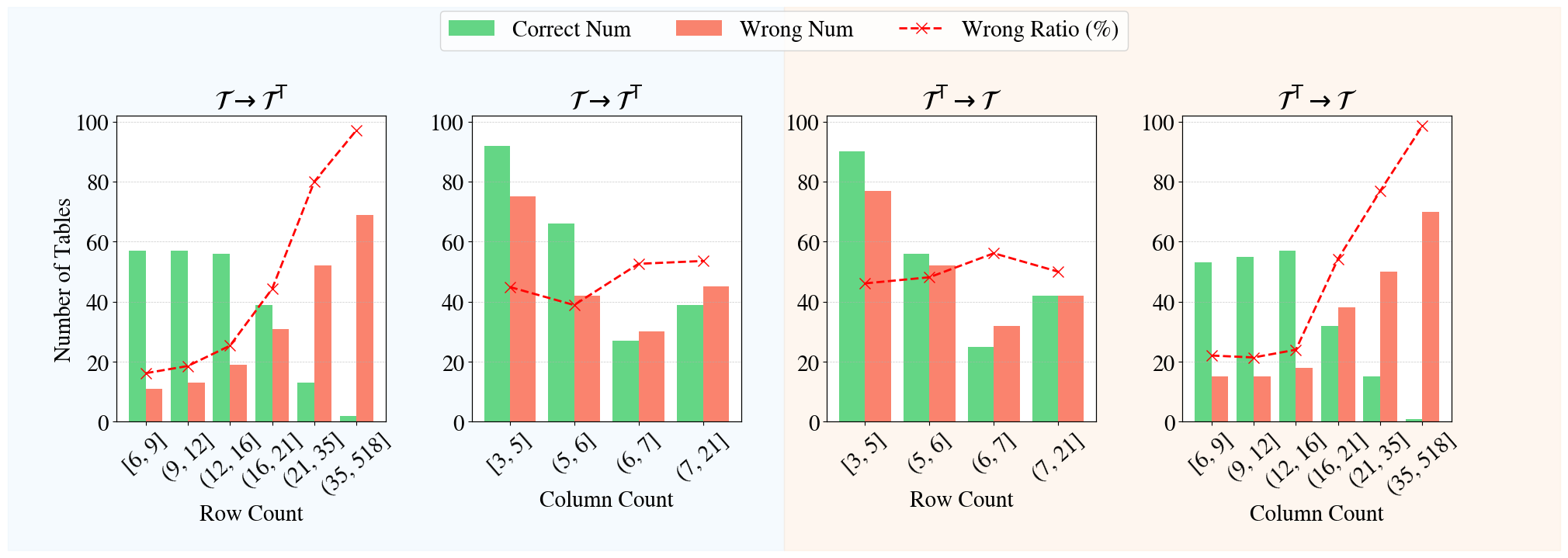}
    \caption{Performance of GPT-3.5 as direct table transposer: \protect\colorbox[HTML]{E6F7FF}{from original to transposed tables ($\mathcal{T} \rightarrow \mathcal{T}^{\top}$)} and \protect\colorbox[HTML]{FFF5E6}{from transposed to original tables ($\mathcal{T}^{\top} \rightarrow \mathcal{T}$)}, with different row and column counts.}
    \label{fig:transpose_length}
\end{figure*}

A further examination of the results, as shown in \Cref{fig:transpose_length}, illustrates that transposition accuracy for LLMs as direct table transposer is associated with the table's dimensions. The accuracy in row-to-column transposition ($\mathcal{T} \rightarrow \mathcal{T}^{\top}$) is distinctly sensitive to the original table's row count, whereas column-to-row transposition ($\mathcal{T}^{\top} \rightarrow \mathcal{T}$) accuracy is similarly related on the number of columns. This observation can be potentially attributed to the inherent characteristics and organizational structure of table data. In most of the row tables, cells within a given column often display homogeneous data types, such as numerical or temporal values. This homogeneity can pose significant challenges for LLMs, as the models might struggle to differentiate between semantically similar cells during the transposition process, thereby leading to potential misalignments and misplacements, particularly as the number of rows increases. Conversely, in those column tables, cells within a row may exhibit similar data types, introducing analogous challenges and potential errors during transposition.

\newpage
\section{Impact of Table Size on \textsc{WTQ} Performance}
\label{sec:table_size_performance}

This section presents an analysis of the impact of table size (quantified by row numbers) on the performance of different methods when applied to the WikiTableQuestions benchmark. Specifically, we examine how the average accuracy of DP, PyAgent, and the combination by applying mix self-consistency is affected by the number of rows in a table. 

\begin{figure}[!t]
    \centering
    \includegraphics[width=\textwidth]{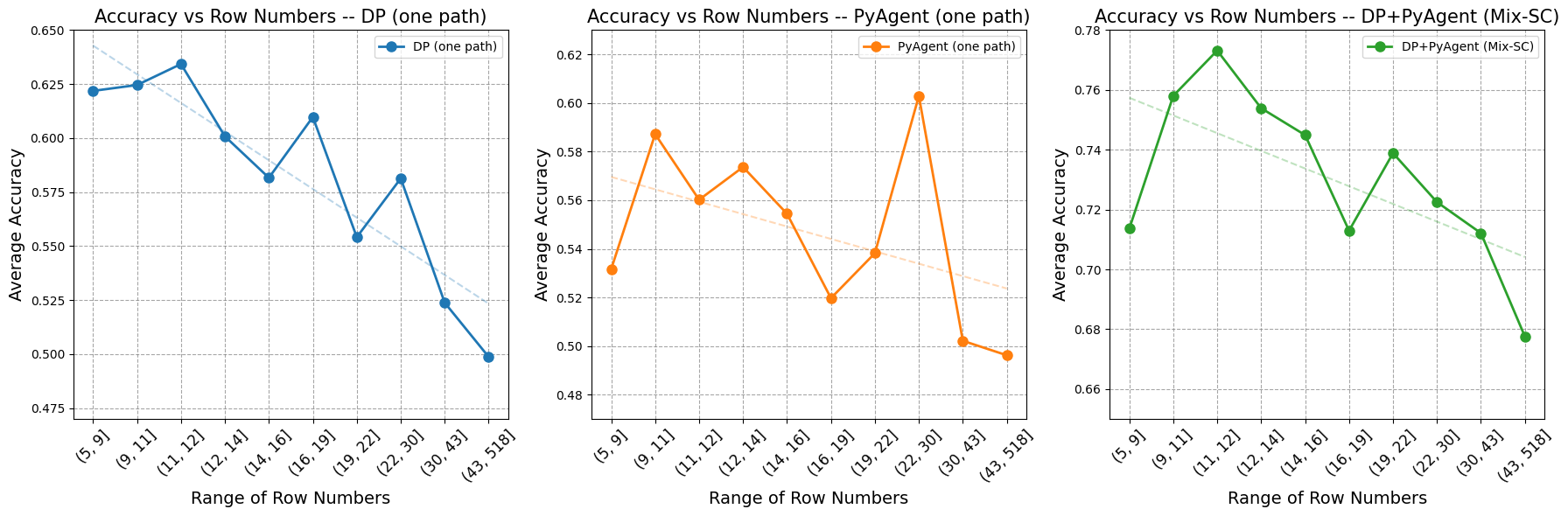}
    \caption{The impact of table size on the accuracy of DP, PyAgent, and Mix-SC on the all the test set of WikiTableQuestions. The x-axis represents the row number ranges, and the y-axis shows the average accuracy for each method.}
    \label{fig:performance_analysis}
\end{figure}

To systematically evaluate the impact, we segmented the row numbers into 10 ranges, each containing approximately 430 data points, and calculated the average accuracy within these intervals. \Cref{fig:performance_analysis} visualizes the average accuracy across these ranges for each method. It is evident that there is a shared trend of diminishing accuracy as the number of rows increases. This observation suggests that all methods are subject to decreased efficacy in the context of long tables.

The decline in performance with larger tables can be attributed to the complexity of handling long-context data and the abundance of potentially interfering information. This complexity often results in an increased error rate. The insights gained from this analysis point towards a need for the development of better symbolic methods for handling long tables, which might be capable of effectively narrowing down the scope of larger tables, either by selective attention to relevant segments or by intelligently summarizing the data, to mitigate the challenges posed by long-context information.

\newpage
\section{Error Case Study for \textsc{WTQ}}
\label{appendix/error_dp_pyagent}
\subsection{Table Misinterpretation}

\subsubsection{Counting Error}
\label{counting_error}

\begin{figure}[!h]
    \centering
    \includegraphics[width=\textwidth]{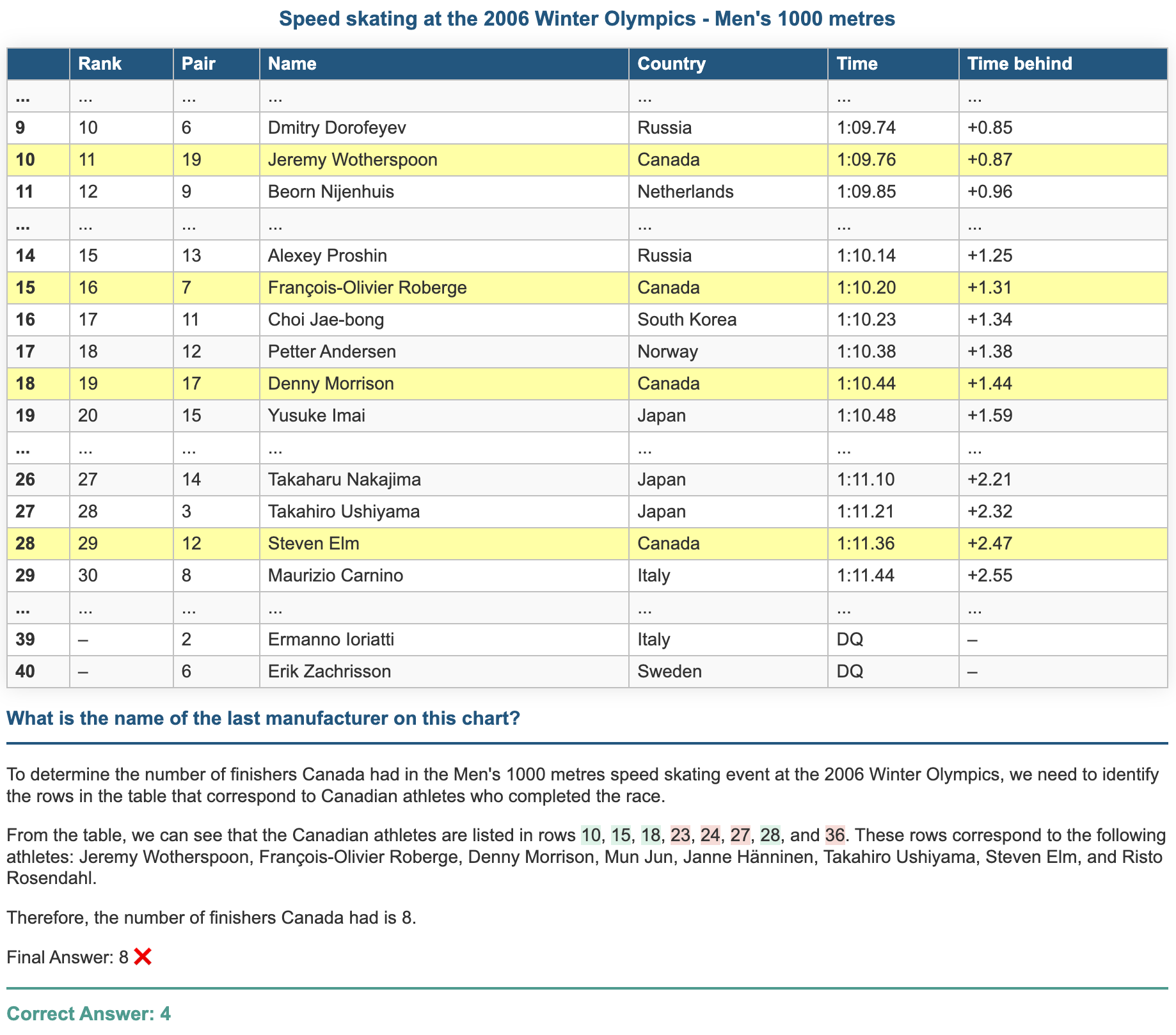}
    \caption{Example of a \textit{table misinterpretation} error by DP, where the LLMs make mistakes attributable to its deficiency in performing counting tasks effectively.}
    \label{fig:error_count}
\end{figure}

\Cref{fig:error_count} highlights a typical error related to table comprehension, emphasizing a common problem in LLMs when dealing with tasks that involve statistical analysis. It points to a weakness in LLMs' ability to accurately process and respond to questions based on statistical information without the help of external tools.

\newpage
\subsubsection{Locating Error}
\label{locating_error}

\begin{figure}[!h]
    \centering
    \includegraphics[width=\textwidth]{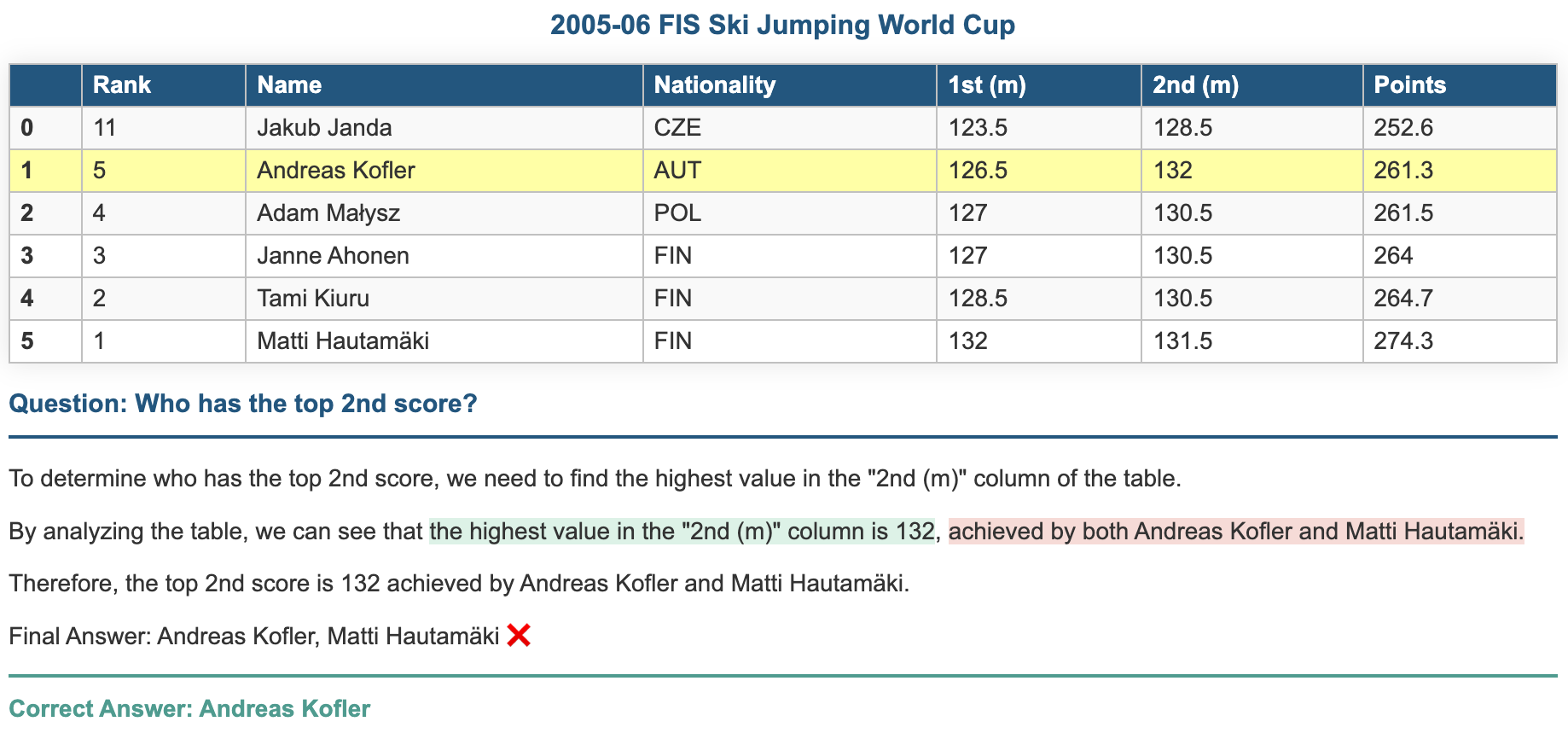}
    \caption{Example of a \textit{table misinterpretation} error by DP, where the model fails accurately locating the specific cell.}
    \label{fig:error_locating}
\end{figure}

\Cref{fig:error_locating} showcases a common error in table interpretation associated with LLMs. This error originates from the LLMs' linearization process, which impairs their ability to recognize table structures. Although the model efficiently identifies the highest value, \texttt{132}, in the \texttt{2nd(m)} column, it inaccurately associates this value with the \texttt{1nd(m)} column, assuming it represents the same feature as in the \texttt{2nd(m)} column. This leads to a misplacement of the value in the table's interpretation.

\newpage
\subsection{Coding Error}

\subsubsection{Attribute Noise Error}
\label{noise_error}

\begin{figure}[!h]
    \centering
    \includegraphics[width=\textwidth]{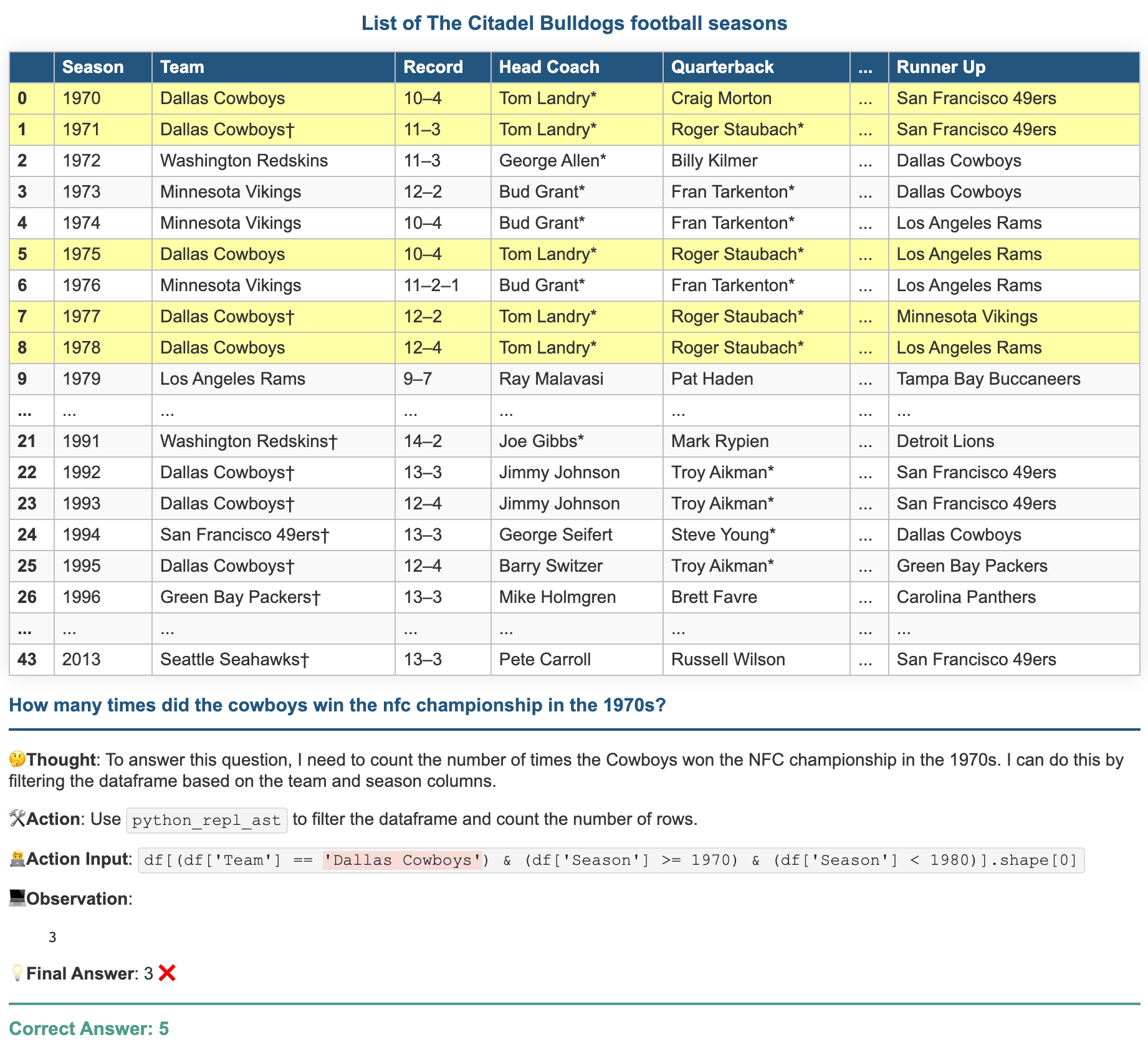}
    \caption{Example of a \textit{Coding Issue} error by PyAgent, where the abnormal variant data entries leads to inaccurate output.}
    \label{fig:error_attribute_noise}
\end{figure}

\Cref{fig:error_attribute_noise} illustrates a minor issue stemming from a coding mistake. The table shown in the figure features a variety of data under the \texttt{Team} column. Alongside the expected \texttt{Dallas Cowboy} entries, there are cells with a slight variation: \texttt{Dallas Cowboy$^\dagger$}. The Python Shell Agent used failed to recognize these unusual variations. This is evident from the use of the \texttt{df[`Team'] = ``Dallas Cowboy''} command for calculating occurrences, leading to a discrepancy in the final count and resulting in inaccurate outcomes.

\newpage
\subsubsection{Special Row Misinterpretation Error}
\label{special_row_error}

\begin{figure}[!h]
    \centering
    \includegraphics[width=\textwidth]{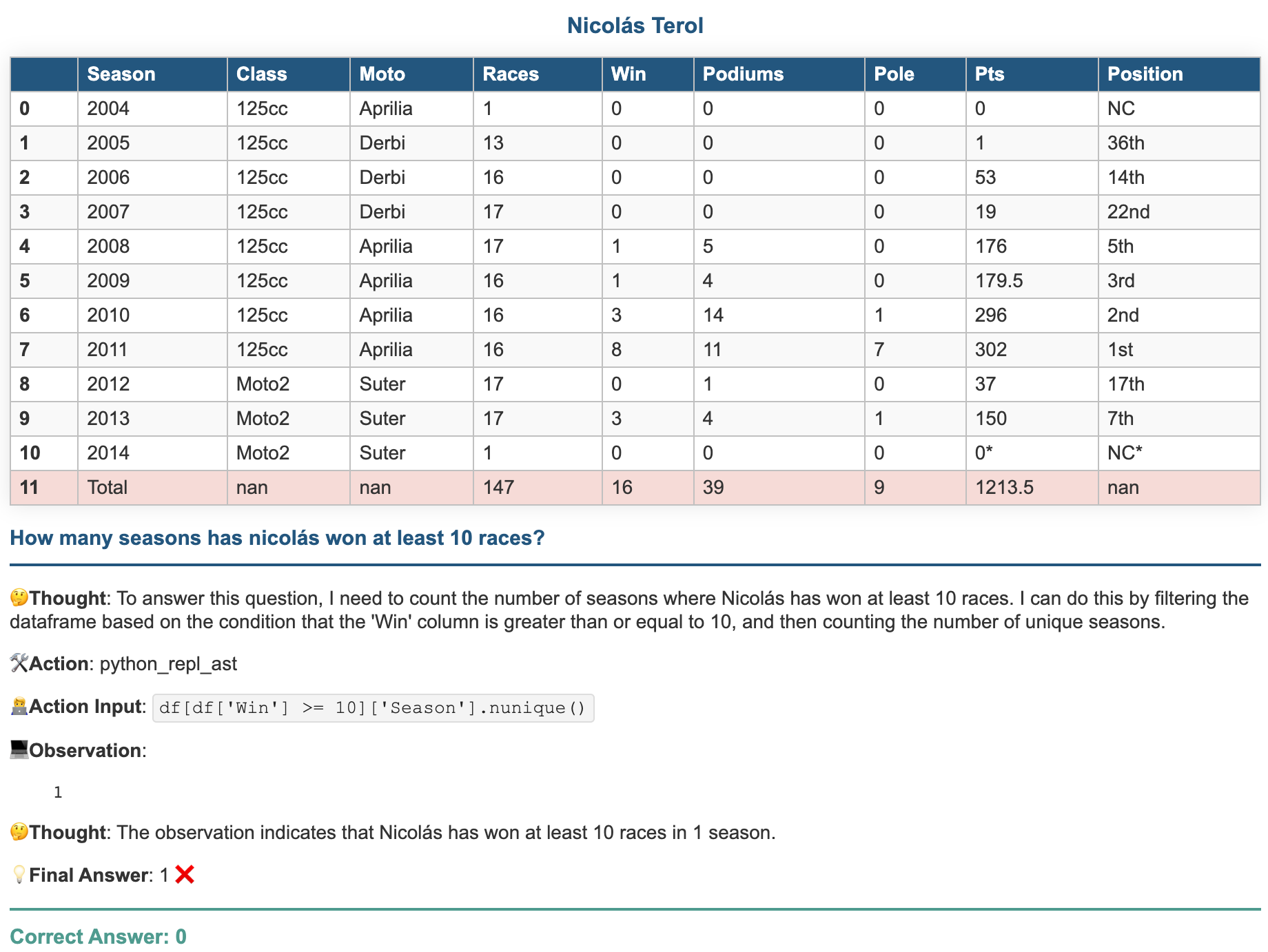}
    \caption{Example of a \textit{coding error} by PyAgent, where PyAgent misinterpretes the special row -- \texttt{Total}.}
    \label{fig:error_special_row}
\end{figure}

\Cref{fig:error_special_row} displays an error where the Python Shell Agent incorrectly interprets a special row, treating the \texttt{Total} row as a valid season entry. In this case, while calculating the seasons in which Nicolás won at least 10 races, the Agent did not properly exclude the cumulative \texttt{Total} row in its code for computation. Consequently, it erroneously reported one season as meeting the criteria.

\newpage
\begin{figure}[!h]
    \centering
    \includegraphics[width=\textwidth]{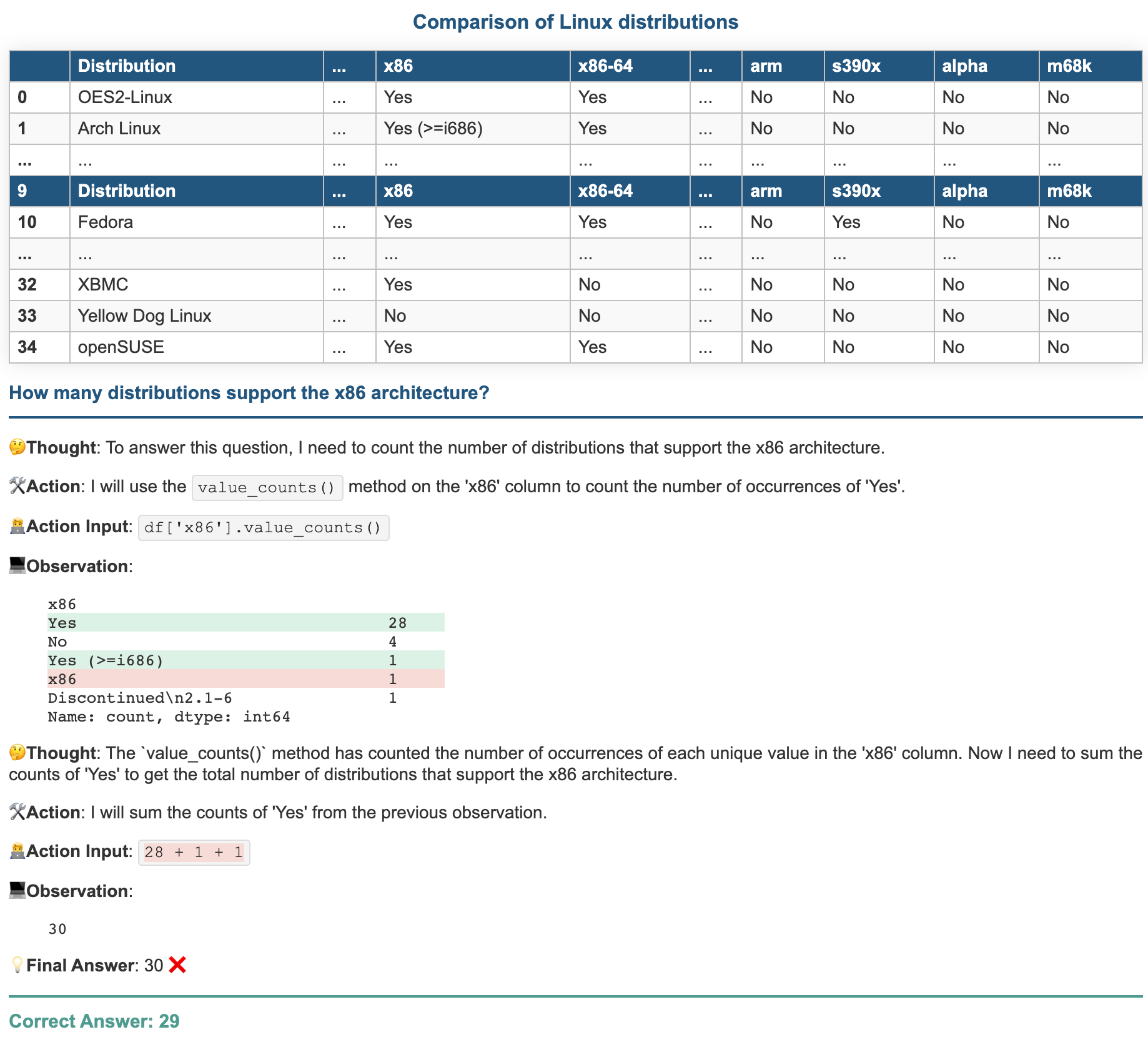}
    \caption{Example of another \textit{coding error} by PyAgent, where the PyAgent misinterpretes the special row which is a nested heading.}
    \label{fig:error_complex}
\end{figure}

\Cref{fig:error_complex} depicts an error where the Python Shell Agent incorrectly includes a special row in its calculations. Specifically, when counting the number of Linux distributions supporting the \texttt{x86} architecture, the agent erroneously counts a nested heading row. As indicated in the figure, the row indexed at 9 is not a valid data entry but rather serves as a nested heading for the table. This row should have been excluded from the count, resulting in an inaccurate calculation ($29 \rightarrow 30$) of distributions supporting the x86 architecture.

\newpage
\subsubsection{Incorrect Coding}
\label{wrong_coding}

\begin{figure}[!h]
    \centering
    \includegraphics[width=\textwidth]{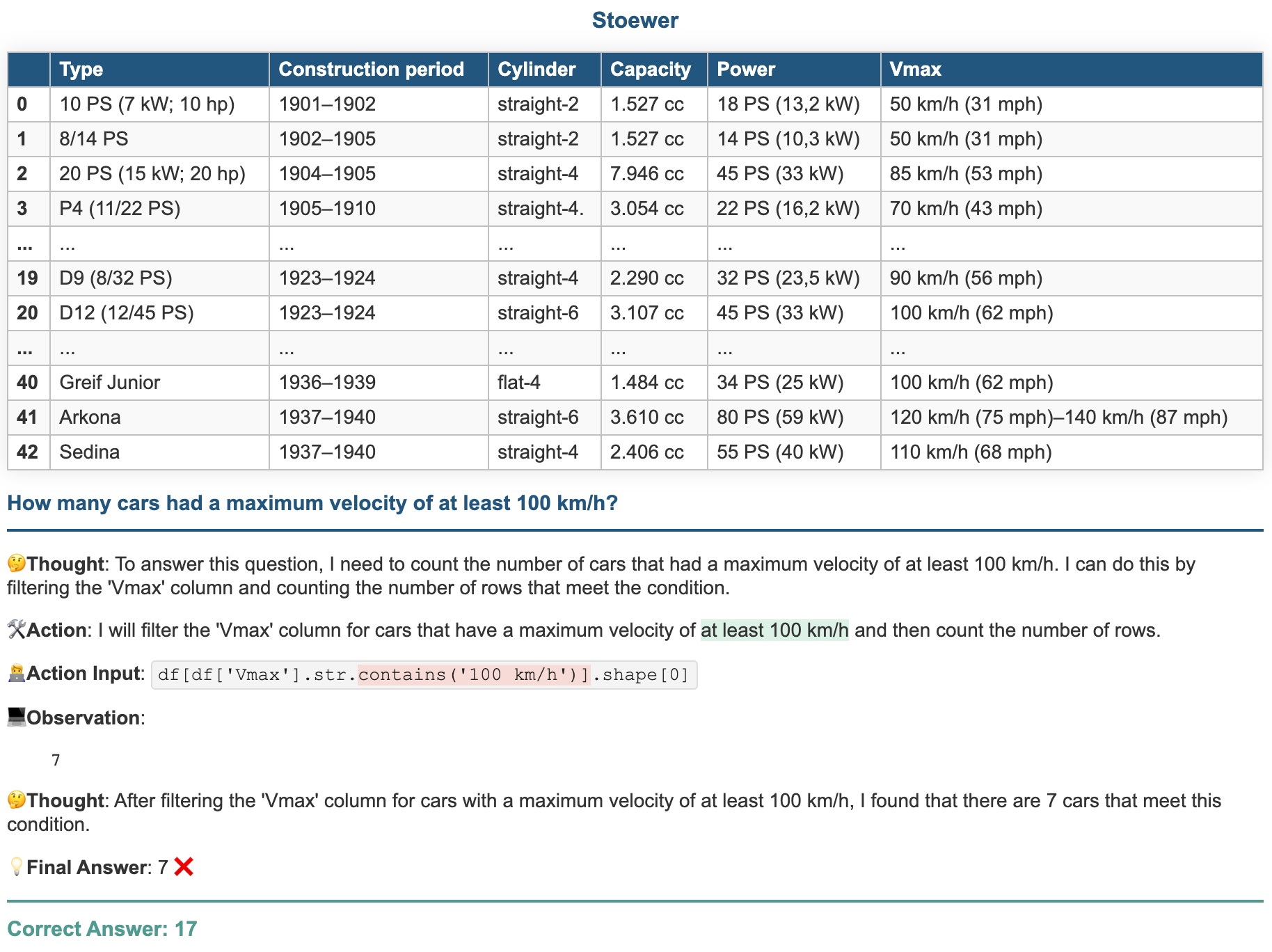}
    \caption{Example of a \textit{coding error} by PyAgent, where the coding is incorrect}
    \label{fig:error_code}
\end{figure}

\Cref{fig:error_code} shows an issue that originates from a basic coding mistake. Although the agent correctly grasps the concept, aiming to filter for entries with \texttt{a maximum velocity of at least 100 km/h}, it falters in the implementation stage by using a \texttt{contains(``100 km/h")} statement in the code. This error is akin to the \textit{Logical Inconsistency} error described in (DP), where a mismatch between the understanding of a concept and its practical execution becomes evident.

\newpage
\subsection{Misalignment Issue}
\subsubsection{Answer Format Issue}
\label{format_error}

\begin{figure}[!h]
    \centering
    \includegraphics[width=\textwidth]{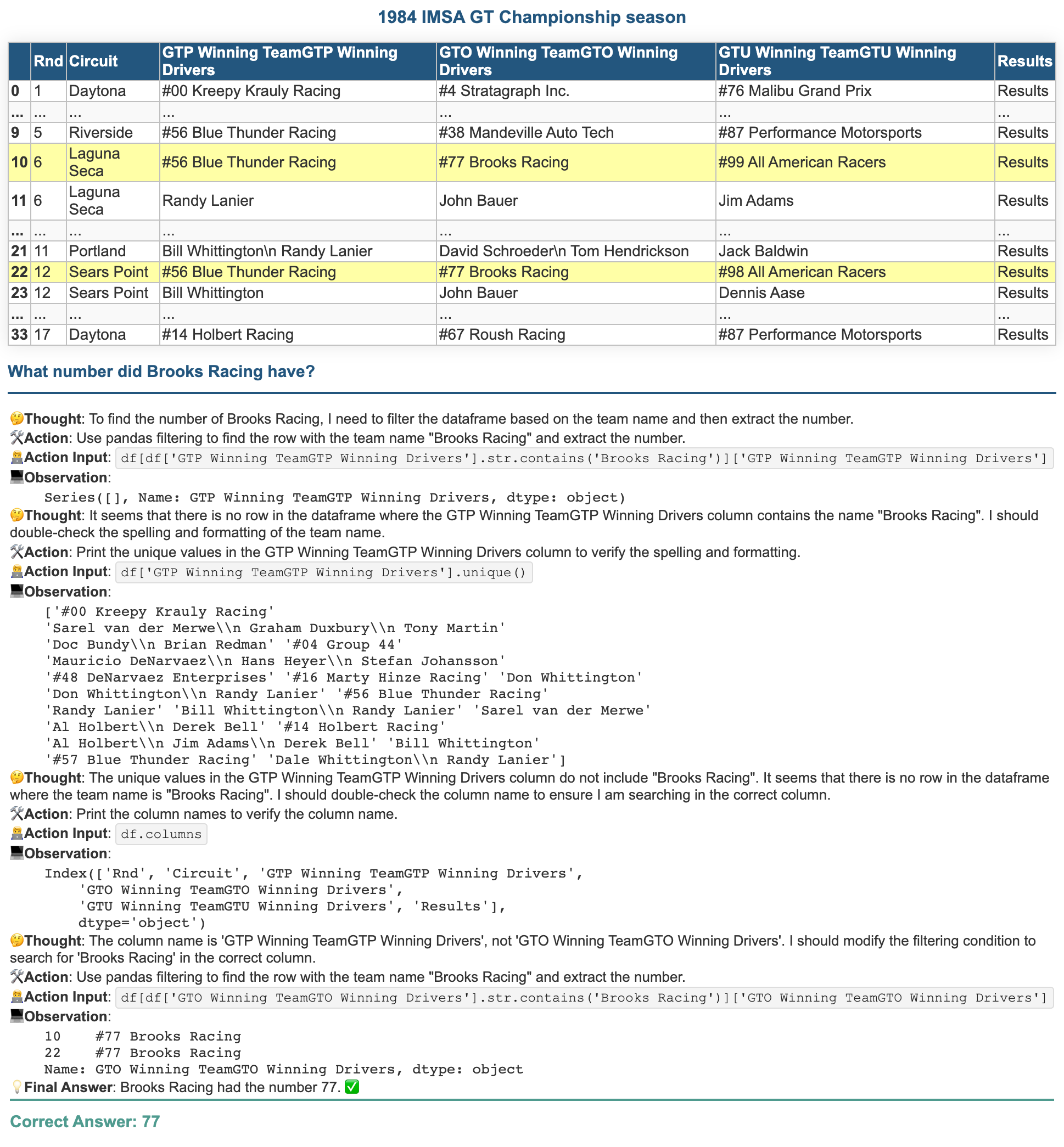}
    \caption{Example of a \textit{misalignment issue} by PyAgent, where the answer format does not follow the instruction in the prompt to give a parsable answer. Note that emojis presented in the figure are purely for visual aid and were not incorporated in actual experiments; the same applies to any figures below.}
    \label{fig:error_answer_parsing}
\end{figure}

\Cref{fig:error_answer_parsing} displays a scenario in which the final answer produced did not align with the specified prompt instructions. In this instance, the Python Shell Agent correctly executed a series of interactions and accurately identified the final answer as \texttt{77}. However, the response given was \texttt{Brooks Racing had the number 77.}. Although the reasoning and the result are correct within their respective contexts, the format of the response impedes the parsing of the correct answer.

\newpage
\subsubsection{Answer Deviation Error}
\label{deviation_error}
\begin{figure}[!h]
    \centering
    \includegraphics[width=\textwidth]{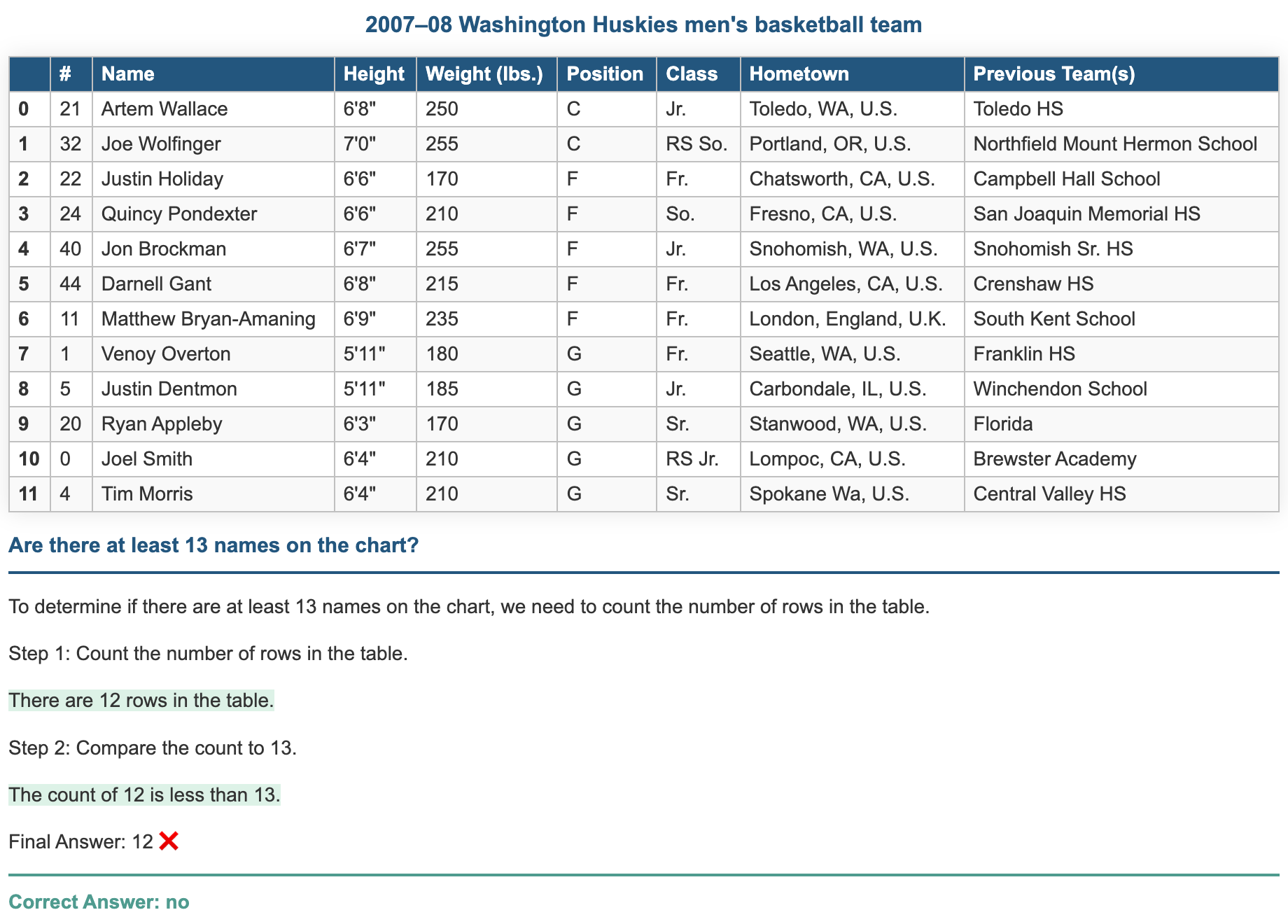}
    \caption{Example of a \textit{misalignment issue} by DP, where the final answer does not directly answer the question.}
    \label{fig:error_answer_off_dp}
\end{figure}

\Cref{fig:error_answer_off_dp} shows a DP example where the final answer does not directly address the posed question. Faced with the binary query \texttt{Are there at least 13 names on the chart?}, the anticipated response should be a simple \texttt{yes} or \texttt{no}. However, the LLM responds with \texttt{12}. Although this answer aligns conceptually with the data in the table and the logic seems accurate, it does not conform to the direct and binary response format required by the question.

\newpage
\begin{figure}[!h]
    \centering
    \includegraphics[width=\textwidth]{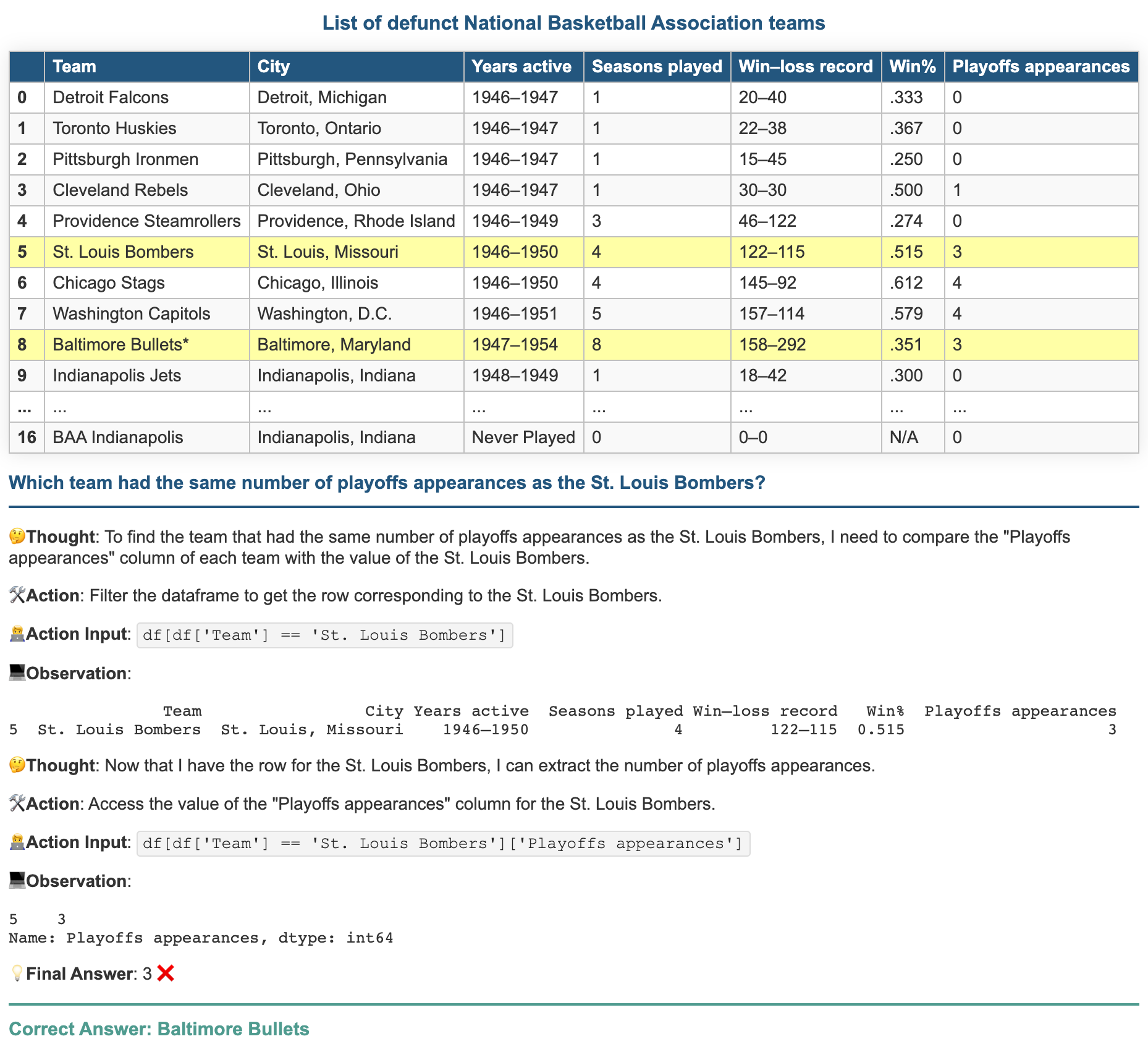}
    \caption{Example of a \textit{misalignment issue} by PyAgent, where the final answer does not directly answer the question.}
    \label{fig:error_answer_off_pyagent}
\end{figure}

\Cref{fig:error_answer_off_pyagent} illustrates a case with the Python Shell Agent where the final response fails to directly answer the posed question. The question \texttt{Which team has the same number of playoffs appearances as the St. Louis Bombers?} clearly requests the identification of a specific team. Yet, it is noted that the agent prematurely delivers an answer upon finding data related to the playoff appearances of the St. Louis Bombers. While the direction of the python shell agent's reasoning appears correct, the resultant answer ultimately falls short of resolving the question correctly.

\newpage
\subsection{Logical Inconsistency}
\label{reasoning_error}
\subsubsection{Reasoning Conflict in DP}
\label{reasoning_error_dp}

\begin{figure}[!h]
    \centering
    \includegraphics[width=\textwidth]{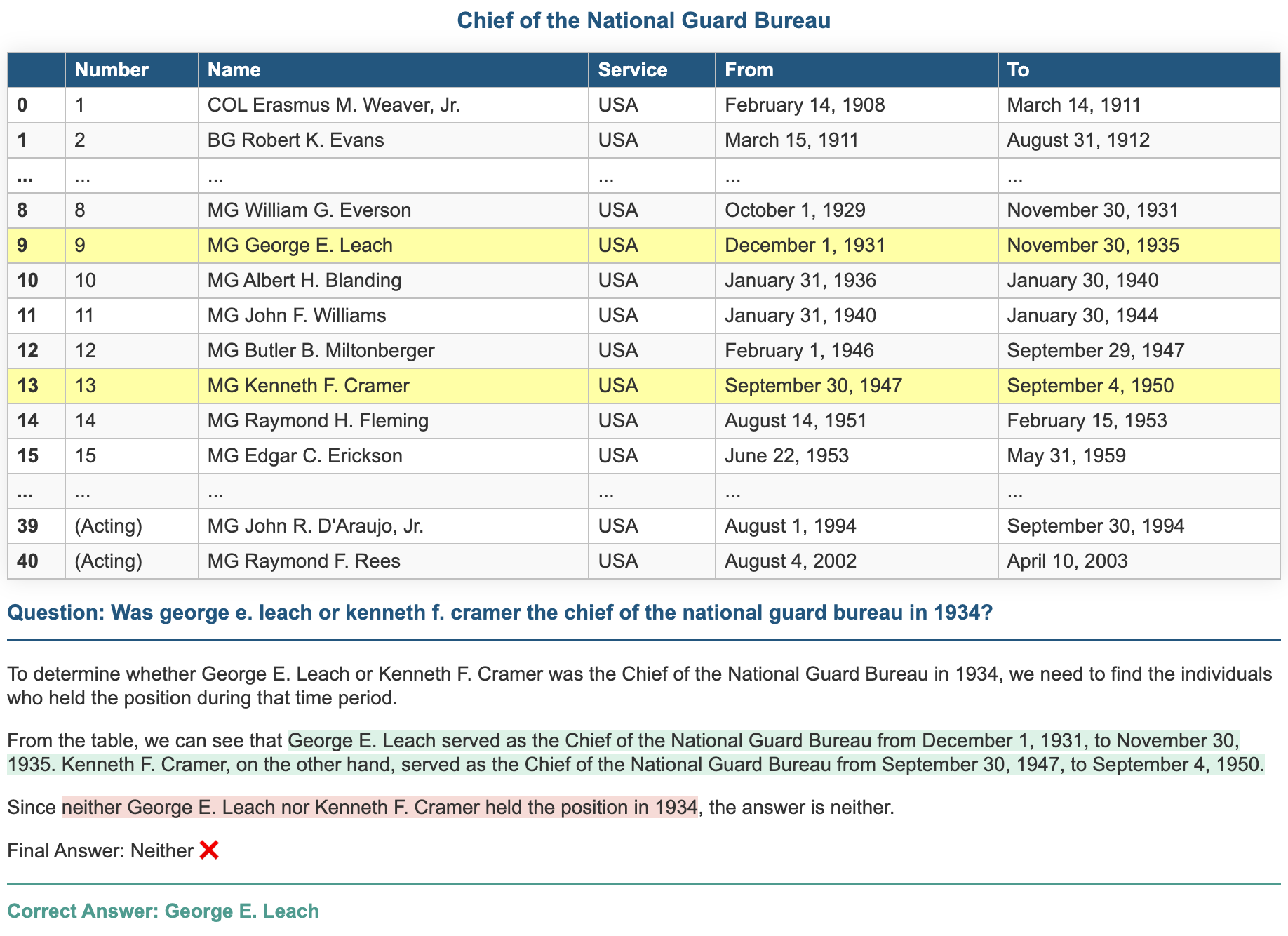}
    \caption{Example of a \textit{logical inconsistency} by DP where a problem with logical reasoning leads to a reasoning conflict in the context.}
    \label{fig:error_reasoning}
\end{figure}

\Cref{fig:error_reasoning} presents an example of a \textit{Logical Inconsistency} error occurring during the interpretation of tabulated data. The error in reasoning is occurred in determining whether George E. Leach or Kenneth F. Cramer was the Chief of the National Guard Bureau in 1934. The reasoning text accurately states that \texttt{George E. Leach served from December 1, 1931, to November 30, 1935}, and \texttt{Kenneth F. Cramer served from September 30, 1947, to September 4, 1950}. Despite this, the interpretation erroneously concludes that neither was in the role in 1934, leading to a contradiction between the information and the final answer.

\newpage
\subsubsection{Reasoning Mistakes in PyAgent}
\label{reasoning_error_pyagent}

\begin{figure}[!h]
    \centering
    \includegraphics[width=\textwidth]{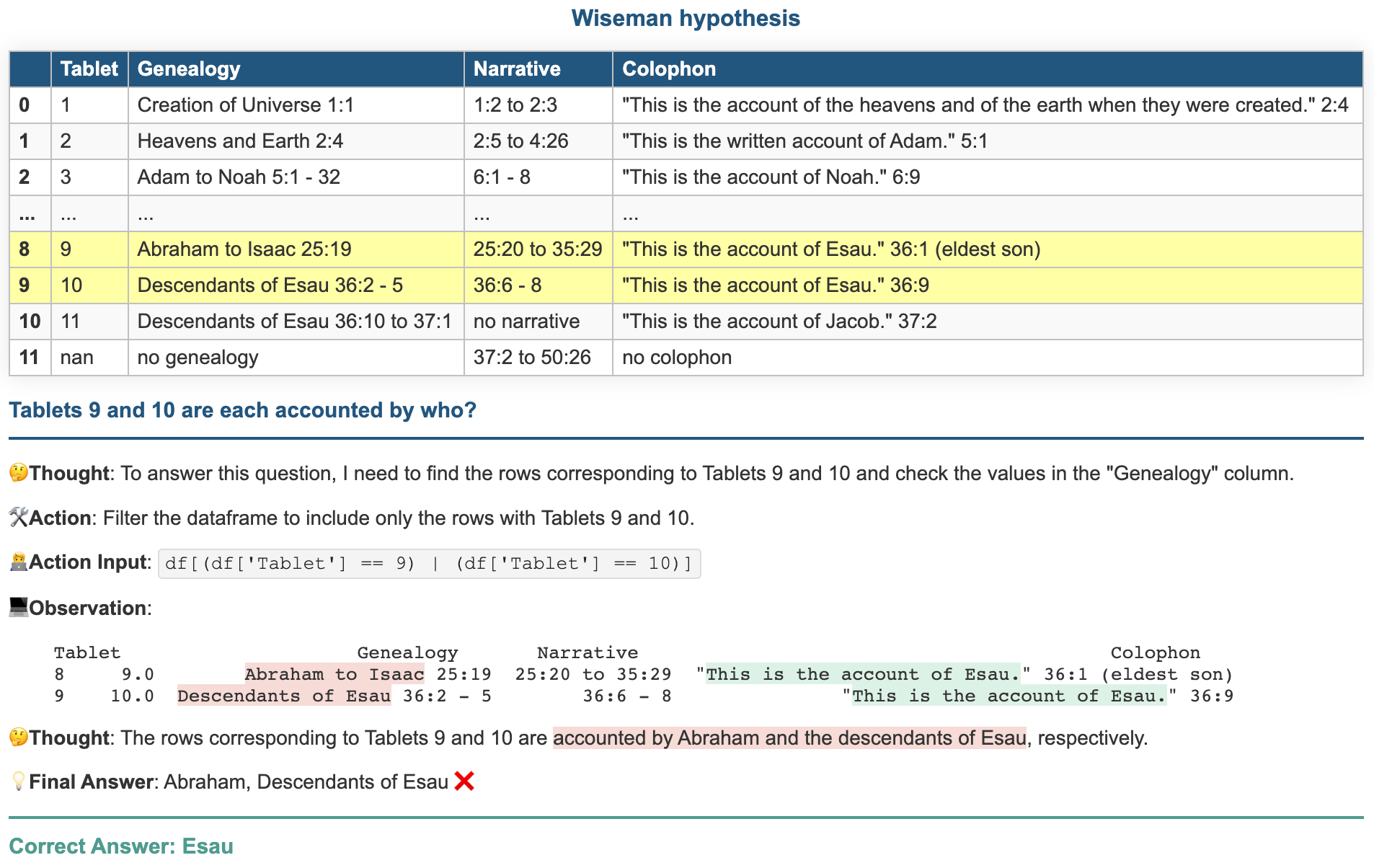}
    \caption{Example of a \textit{logical inconsistency} error by PyAgent where the LLMs make mistakes in reasoning on the observations.}
    \label{fig:observation}
\end{figure}

\Cref{fig:observation} depicts an instance of a \textit{Logical Inconsistency Error} during the Python Shell interaction. In this case, LLMs successfully the necessary steps to gather information about \texttt{Tablet 9} and \texttt{Tablet 10}, a misinterpretation of the retrieved data results in flawed reasoning. This leads to an error in the conclusion drawn from the interaction.

\newpage
\subsection{Execution Issue}
\subsubsection{Interaction Bound or Looping Error}
\label{loop_error}

\begin{figure}[!h]
    \centering
    \includegraphics[width=\textwidth]{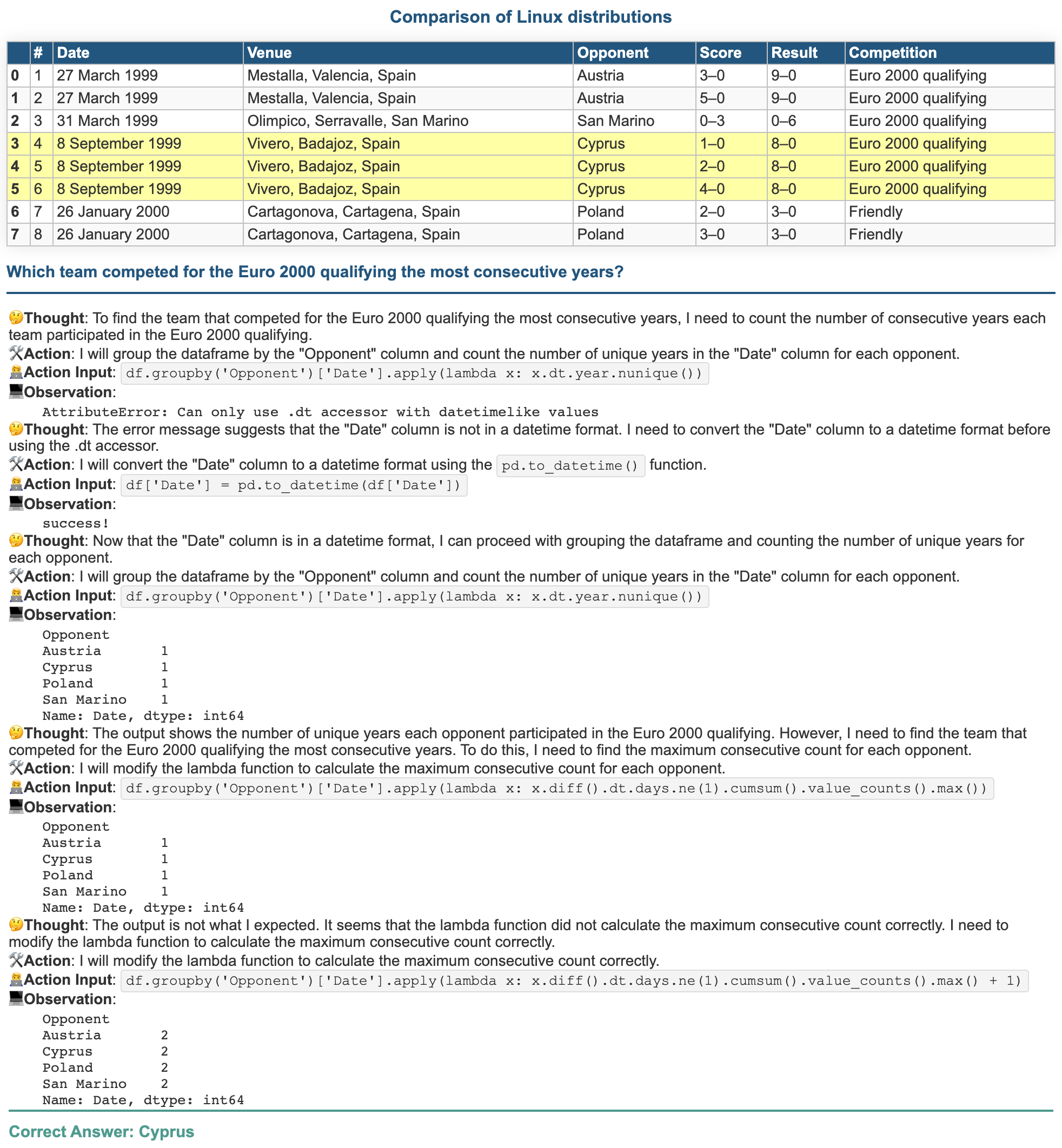}
    \caption{An example of an \textit{execution issue} by PyAgent, where the agent attempts to fix an coding error but falls into a loop.}
    \label{fig:error_loop}
\end{figure}

\Cref{fig:error_loop} demonstrates an instance of \textit{Execution Issue} error made by the Python Shell Agent. In the process of identifying the team that participated in Euro 2000 qualifying for the most consecutive years, the agent faces difficulties in the data processing phase. Initially, an error occurs due to the \texttt{Date} column not containing \texttt{datetime} objects. Then the agent successfully converts the entries into the appropriate format. However, the agent, in trying to compute the number of consecutive participation years for each team, gets stuck in a loop of continually refining its calculation method without arriving at a conclusive answer within the given interaction steps. 

\newpage
\subsubsection{Non-Observable Action Error}
\label{non_observable_error}

\begin{figure}[!h]
    \centering
    \includegraphics[width=\textwidth]{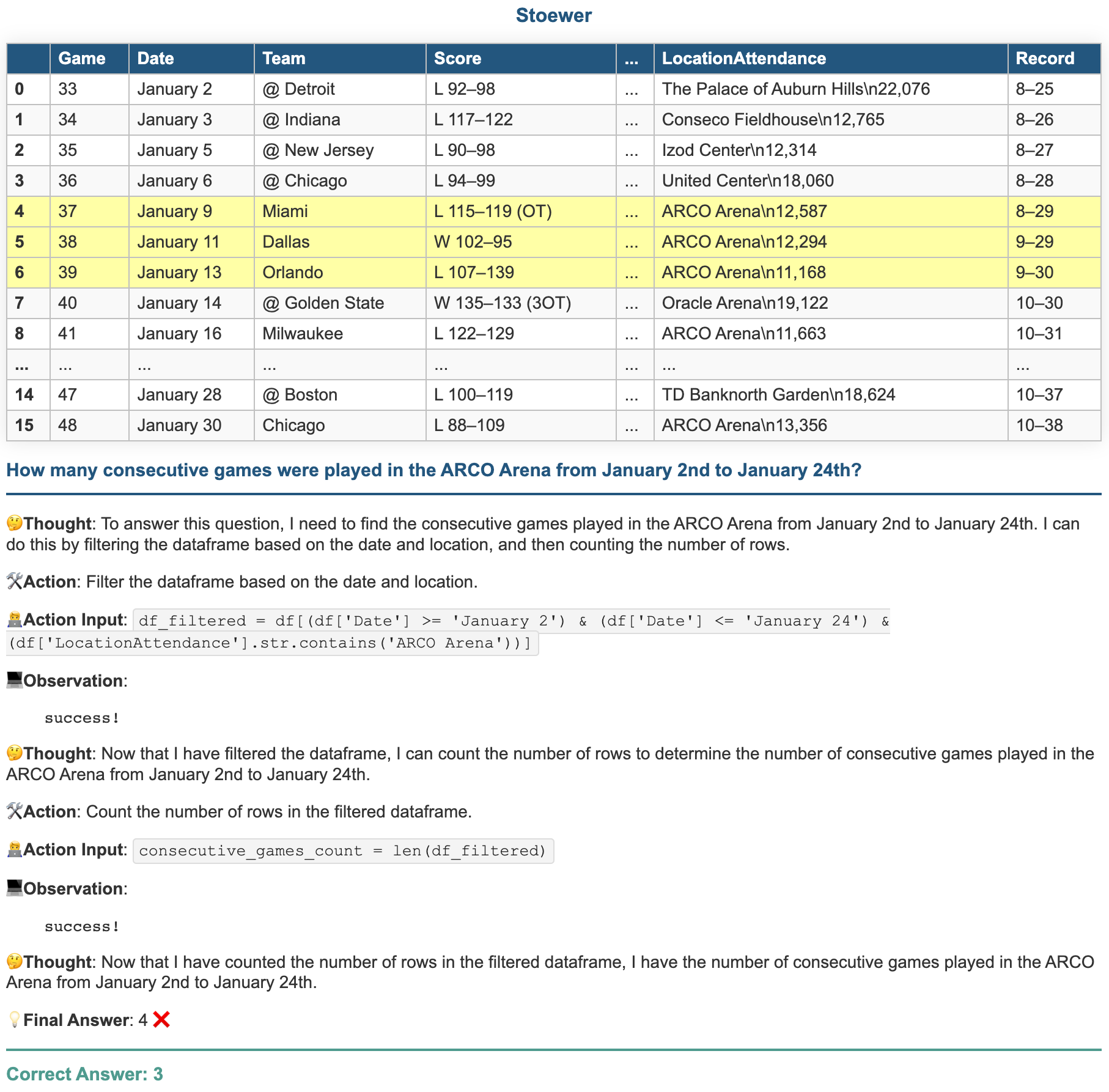}
    \caption{Example of a \textit{Execution Issue} error by PyAgent, where the agent inputs non-observable actions into Python Shell.}
    \label{fig:error_noprint}
\end{figure}

\Cref{fig:error_noprint} illustrates a situation where PyAgent encounters a \textit{Non-Observable Action Error}. Specifically, the actions given by the PyAgent, for example, \texttt{consecutive\_games\_count = len(df\_filtered)}, do not generate any observable output in the Python Shell. Consequently, the agent does not receive any valid observations to aid it in deriving the correct answer, thus, leading to an incorrect count of consecutive games played in the ARCO Arena from January 2nd to January 24th.

\newpage
\subsection{Resorting Issue}
\label{norm_error}

\begin{figure}[!h]
    \centering
    \includegraphics[width=\textwidth]{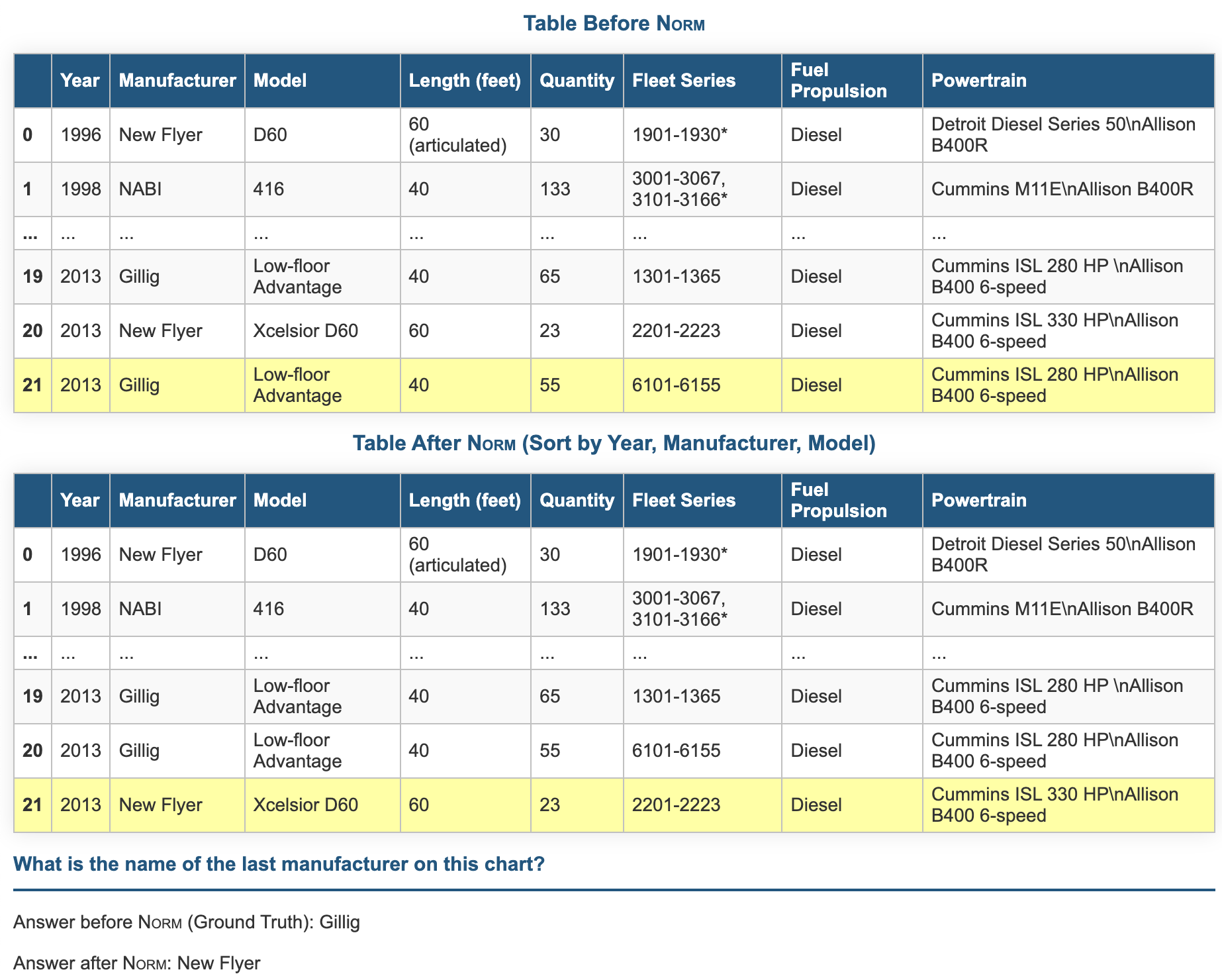}
    \caption{Example of a \textit{Normalization Issue} error by DP, where the correct answer changes due to the resorting stage in \textsc{Norm}.}
    \label{fig:error_norm}
\end{figure}

\Cref{fig:error_norm} shows a case of data inconsistency due to the application of the resorting stage in the \textsc{Norm} procedures. The figure's upper table displays the original format, with \texttt{Gillig} as the manufacturer in the final row. However, after resorting as suggested by LLMs, the lower table in the figure lists \texttt{New Flyer} as the last row's manufacturer. This change, while seemingly minor in the broader context of table comprehension, significantly impacts responses to specific queries like ``\texttt{What is the name of the last manufacturer on the chart?}''

\newpage
\section{Analysis of Mix Self-Consistency}
\label{analysis_MSC}

\subsection{Ablation Study of Output Selection}
\label{ablation_output_selection}

This section presents an ablation study conducted to elucidate the effect of various combinations of DP and PyAgent outputs on the performance of the \textit{Mix Self-Consistency} method. For this experiment, we systematically explored different combinations while keeping the total output count constant at ten. Each combination was tested 100 times through random shuffling. For each test, maximum, minimum, and average accuracies were recorded.

\begin{figure}
    \centering
    \includegraphics[width=\textwidth]{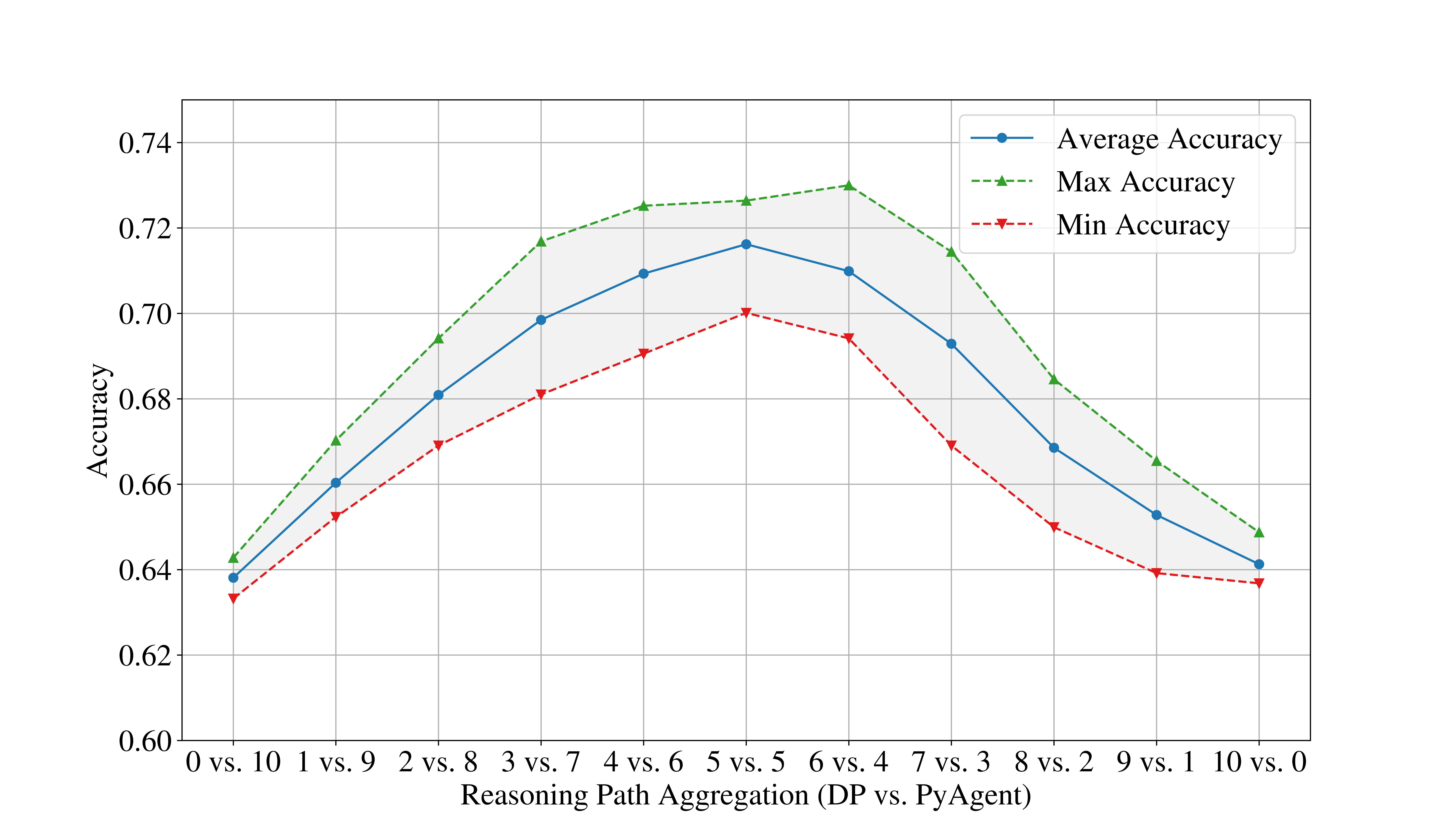}
    \caption{Accuracy results for the \textit{Mix Self-Consistency} method applied to the sampled \textsc{WTQ} dataset, with varying combinations of DP and PyAgent outputs (depicted as DP vs. PyAgent on the x-axis). The combinations range from 10 DP vs. 0 PyAgent to 0 DP vs. 10 PyAgent. Each data point represents the maximum, minimum, and average accuracies obtained from 100 tests per combination, conducted using random sampling. Note that for the 10 DP vs. 0 PyAgent and 0 DP vs. 10 PyAgent combinations, there is no random sampling of paths. However, variance is observed due to the presence of multiple equally probable answer sets generated by the 10 paths, leading to different possible selections of answers even without sampling, thereby introducing randomness into the results.}
    \label{fig:ablation_study}
\end{figure}

\Cref{fig:ablation_study} shows the results of the ablation study. The 5+5 combination (5 DP + 5 PyAgent) consistently gives the highest minimum and average accuracies among all tested combinations, making it a robust and reliable choice for this task. The 4+6 combination (4 DP + 6 PyAgent) secured the highest maximum accuracy in our tests.

Through this ablation study, we aim to provide insights into how different output selections influence the effectiveness of the \textit{Mix Self-Consistency} method. Importantly, the choice of output combination should be considered as a hyperparameter that is intimately related to the distribution of the dataset being used. Given that different reasoning strategies exhibit unique strengths and weaknesses, it is crucial to tailor the output combination to align with the characteristics of the specific tasks and datasets in question, thereby maximizing the performance of the \textit{Mix Self-Consistency} method.

\subsection{Mechanics of Mix Self-Consistency in Output Selection}
\label{mechanics_of_msc}

\begin{figure}[!t]
    \centering
    \includegraphics[width=0.8\linewidth]{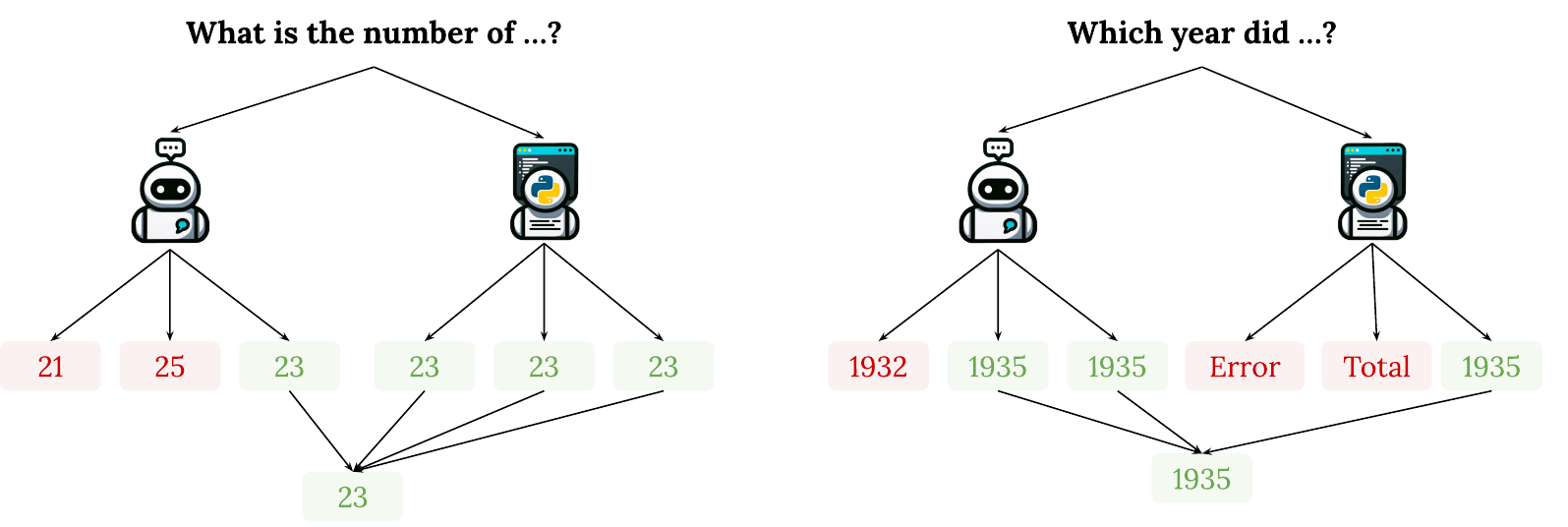}
    \caption{An illustration of \textit{Mix Self-Consistency} by aggreagting outputs from multiple reasoning methods to form a unified, high-confidence prediction..}
    \label{fig:msc}
    \vspace{-1.5em}
\end{figure}

The effectiveness of the \textit{Mix Self-Consistency} method in achieving high accuracy largely stems from its ability to harness the strengths of different reasoning methods. Intuitively, the multiple outputs from certain reasoning method can be interpreted as the confidence score for the generated answers. In scenarios where a method excels, its outputs often tend to converge towards a common answer, signifying higher confidence and reliability. In contrast, a method less suited to the problem at hand tends to produce more diverse results, indicative of a lower level of confidence. By aggregating these outputs from different methods and applying majority voting, the \textit{Mix Self-Consistency} method refines these variations into a more accurate prediction. As shown in \Cref{fig:msc}, This process leverages the strengths of the employed reasoning methods, thereby enhancing overall performance.

\section{Results of Mix Self-Consistency on TabFact}
\label{results_tabfact}

This section presents the additional results of applying the \textit{Mix Self-Consistency} method to the TabFact dataset, as part of an extended investigation to verify and evaluate the method's adaptability and effectiveness in other related tasks beyond WTQ dataset.

\begin{wraptable}{r}{0.5\textwidth}
\begin{tabular}{lc}
\toprule
\textbf{Method} & \textbf{Accuracy} \\
\midrule
StructGPT~\cite{Jiang-StructGPT-2022} & 0.708 \\
Dater~\cite{dater} & 0.874 \\
\midrule
\textbf{Ours} &\textbf{ 0.885} \\
\toprule
\end{tabular}
\caption{Accuracy results of different methods without fine-tuning on the TabFact dataset.}
\label{tab:accuracy_results}
\end{wraptable}

For TabFact, a subsample of 500 data points was randomly selected from the test set. The experimental setup mirrored that of the WTQ experiments, employing the same parameters such as temperature settings for model inference. The strategy for output selection in the TabFact experiment also follows the 5+5 combination, which proves to be the best for the WTQ dataset, to aggregate the output answers from 5 instances of DP and 5 instances of PyAgent. Additionally, all the prompts (e.g., DP, PyAgent) used in the TabFact experiment were slightly modified to align with the requirements of the fact-checking scenarios.

\Cref{tab:accuracy_results} summarizes the accuracy results of the \textit{Mix Self-Consistency} method, StructGPT, and Dater on the TabFact dataset. \textit{Mix Self-Consistency} can also achieve the highest accuracy, outperforming both StructGPT and Dater in fact-checking.

\end{document}